\RequirePackage[svgnames,table]{xcolor}
\documentclass[11pt,letterpaper]{yalearxiv}

\usepackage{graphicx}
\usepackage{float,epstopdf}
\usepackage{bbm}

\usepackage{microtype}

\usepackage{natbib}
\setcitestyle{square}

\usepackage{subcaption}
\usepackage{booktabs}

\usepackage{cleveref}

\usepackage{amsmath}
\usepackage{amssymb}
\usepackage{mathtools}
\usepackage{amsthm}
\usepackage{dsfont}
\usepackage{multicol}
\usepackage{makecell}
\usepackage{multirow} 
\usepackage{amsfonts} 
\usepackage{mathrsfs}
\usepackage[amssymb, thickqspace]{SIunits}
\usepackage{enumitem}
\usepackage{pgfplotstable}

\usepackage{microtype}
\usepackage{graphicx}
\usepackage{booktabs} % for professional tables
\usepackage[table]{xcolor}
\usepackage{arydshln}
\usepackage[normalem]{ulem} % [normalem] prevents the package from changing the default behavior of `\emph` to underline.

\usepackage{cases}
\usepackage{wrapfig}

\usepackage{algorithm}
\usepackage{algorithmic}

\usepackage{url}

\usepackage{thmtools}
\usepackage{thm-restate}
\usepackage{tabu}

\definecolor{huskypurple}{HTML}{4B2E83}

\usepackage{titletoc}

\usepackage{listings}
\lstdefinestyle{promptstyle}{
  basicstyle=\ttfamily\footnotesize,
  breaklines=true,
  breakautoindent=false,
  breakindent=0pt,
  postbreak=\mbox{\textcolor{gray}{$\hookrightarrow$}\space},
  columns=fullflexible,
  keepspaces=true,
  frame=single,
  framesep=5pt,
  xleftmargin=6pt,
  xrightmargin=6pt,
  aboveskip=8pt,
  belowskip=8pt,
  showstringspaces=false,
}   % third-party packages (natbib, booktabs, listings, ...)

\newcommand{\ghat}{\widehat{g}}

\newcommand{\rhat}{\widehat{r}}

\newcommand{\mbE}{\mathbb{E}}

\newcommand{\mbP}{\mathbb{P}}

\newcommand{\mbR}{\mathbb{R}}

\newcommand{\Ical}{\mathcal{I}}

\newcommand{\Scal}{\mathcal{S}}

\newcommand{\Xcal}{\mathcal{X}}
\newcommand{\Ycal}{\mathcal{Y}}

\newcommand{\argmax}{\mathop{\mathrm{argmax}}}

\newcommand{\supp}{\mathop{\mathrm{supp}}}

\newcommand{\rstar}{r^\star}
\newcommand{\gstar}{g^\star}

\newcommand{\piref}{\pi_\mathrm{ref}}

\newcommand{\bon}{Bo$N$}
\newcommand{\cbon}{CBo$N$}

\newcommand{\ind}{\mathbb{I}}
\newcommand{\relu}{\operatorname{ReLU}}

\newcommand{\epsr}{\varepsilon_r}
\newcommand{\epsg}{\varepsilon_g}
\newcommand{\safehat}{\widehat{\Scal}_+}

\newcommand{\cPes}{\textsf{cPes}}

\newtheorem{definition}{Definition}[section]
\newtheorem{theorem}[definition]{Theorem}

\newtheorem{remark}[definition]{Remark}
\newtheorem{corollary}[definition]{Corollary}
\newtheorem{proposition}[definition]{Proposition}

\newtheorem{assumption}[definition]{Assumption}

\newtcolorbox{simpleElegantQuote}{
    colback=AliceBlue!50!White,
    colframe=RoyalBlue!75!Black,
    boxrule=0.5pt,
    arc=2mm,
    boxsep=4pt,
    left=10pt, right=10pt,
    top=8pt, bottom=8pt,
    fontupper=\itshape,
}

\title{Safety Hacking in Constrained Best-of-$N$ \\ Inference-time Scaling}

\runningtitle{Safety Hacking in Constrained Best-of-$N$ Inference-time Scaling}

\usepackage{fontawesome5}   % brand/utility icons, e.g. \faGithub

\definecolor{yaleblue}{RGB}{0,34,195}
\newcommand{\ly}{%
  \textsuperscript{%
    {\usefont{T1}{pbk}{m}{n}\textcolor{yaleblue}{1}}%
  }%
}

\newcommand{\utriken}{%
  \textsuperscript{%
    {\usefont{T1}{pbk}{m}{n}\textcolor{yaleblue}{2,3}}%
  }%
}

\newcommand{\ut}{%
  \textsuperscript{%
    {\usefont{T1}{pbk}{m}{n}\textcolor{yaleblue}{2}}%
  }%
}

\newcommand{\riken}{%
  \textsuperscript{%
    {\usefont{T1}{pbk}{m}{n}\textcolor{yaleblue}{3}}%
  }%
}

\author{Akifumi Wachi\ly \quad Takumi Tanabe\ly \quad Youhei Akimoto\utriken \\
\ly LY Corporation \quad \ut University of Tsukuba \quad \riken RIKEN AIP \\
Correspondence: \texttt{akifumi.wachi@lycorp.co.jp} \\
}

\hypersetup{colorlinks=true, linkcolor=blue!50!black, citecolor=blue!50!black,
            urlcolor=blue!50!black}

\begin{document}

% -----------------------------------------------------------------------------
%  Abstract. In this template the abstract is *captured* here and then typeset
%  inside the title box by \maketitle, so it must appear BEFORE \maketitle.
%  Keep it to a single paragraph.
% -----------------------------------------------------------------------------
\begin{abstract}
{\centering\section*{Abstract}}
Inference-time pipelines often sample multiple outputs, filter them with a learned safety model, and return the proxy-feasible output with the highest learned reward. We show that this composition creates a two-stage failure: an imperfect safety proxy first contaminates the feasible set with unsafe outputs, and reward maximization can then amplify this residual contamination. We define \emph{safety hacking} as selecting an output that passes the learned constraint but violates the true safety criterion. For constrained Best-of-$N$ sampling, we derive finite-$N$ bounds governed by the joint upper reward tails of safe and unsafe outputs within the proxy-feasible set. If unsafe-but-feasible outputs have the heavier tail, safety hacking becomes asymptotically certain as $N$ grows, even when false-positive mass and average safety- and reward-proxy errors are arbitrarily small. We also show that policies within a bounded $\chi^2$ divergence from the proxy-feasible reference distribution admit an $N$-independent safety-hacking bound, and instantiate this general coverage-control principle with constrained pessimistic sampling. Coverage control limits amplification but cannot repair a contaminated feasible set: admitted unsafe outputs may still be favored, and regularized selection is not necessarily safer than constrained Best-of-$N$ for every reward proxy. Toy and language-model experiments characterize both contamination and its reward-tail amplification, which exposes an inherent difficulty in inference-time scaling with learned safety models.
\end{abstract}

\maketitle

\section{Introduction}
\label{sec:introduction}

Foundation-model inference increasingly relies on learned models both to enforce safety and to assess response quality \citep{snell2024scaling,amodei2016concrete,hendrycks2021unsolved}.
A typical pipeline samples candidate responses, rejects those deemed unsafe by a safety model, and ranks the remainder with a reward model \citep{wang2025safety,ji2025almost,inan2023llama,zeng2024shieldgemma}.
As a representative instance, constrained Best-of-$N$ (\cbon) sampling follows this pattern: it draws $N$ responses and returns the proxy-feasible response with the largest learned reward.

Filtering and ranking errors can interact.
The safety model may admit a response that violates the true safety criterion, thereby contaminating the proxy-feasible set; the reward model may then assign that response a high score.
We call the resulting selection event \emph{safety hacking}: the returned response satisfies the learned safety constraint but violates the true one.
Such \emph{unsafe-but-feasible} responses remain inside the feasible set and are therefore exposed to downstream reward optimization, which can amplify a small amount of contamination into a large selection probability.

Existing work on reward overoptimization shows that excessive optimization against an imperfect reward proxy can eventually reduce the true reward~\citep{skalse2022defining,gao2023scaling}.
Our setting adds a distinct source of misspecification: the optimization domain is itself defined by an imperfect safety proxy.
The resulting failure is governed by the interaction between 1)~\emph{contamination} that determines which unsafe outputs enter the proxy-feasible set, and 2)~\emph{amplification} that determines how strongly reward optimization favors them.
Our main result shows that amplification depends not only on how often the filter admits unsafe responses, but also on the relative upper tails of learned rewards among safe and unsafe proxy-feasible responses.
When the unsafe group has the heavier upper tail, its maximum eventually dominates under \cbon{}, and the probability of selecting an unsafe response tends to $1$ as $N \to \infty$.
This can occur even as the mass of unsafe false acceptances and average safety- and reward-proxy errors vanish.
Average filter accuracy is therefore insufficient to predict safety under inference-time scaling; the joint tail behavior of the two proxies also matters.

Safety hacking is related to reward overoptimization and \bon{} jailbreaking \citep{gao2023scaling,hughes2024bestof}, but it is not ordinary reward hacking restricted to a benign feasible set.
A main reason for safety hacking is that reward optimization operates on a set that may already contain false acceptances from the safety filter and can favor those errors over other candidates.
The cleanest example is \Cref{cor:small-rmse-amplification}: an unsafe-but-feasible reference mass $\varepsilon$ can coexist with safety- and reward-proxy errors of order $\sqrt{\varepsilon}$, yet a budget $N=O(1/\varepsilon)$ suffices to produce high-probability safety hacking.

Pessimistic inference can provide scaling-monotonic guarantees for reward overoptimization under suitable conditions \citep{huang2025best}, but it cannot determine which responses were wrongly admitted by a safety proxy.
We show that any policy close to the proxy-feasible reference distribution in $\chi^2$ divergence admits an $N$-independent safety-hacking bound.
Our concrete example is constrained pessimistic sampling (\cPes), a $\chi^2$-regularized reweighting of the proxy-feasible reference distribution.
The guarantee limits concentration on residual errors but cannot eliminate them. 
Depending on the reward proxy, regularized selection may still exhibit substantial safety hacking.

Our contributions are:
\begin{enumerate}
\item We formalize \emph{safety hacking} as selection of a response that
satisfies a learned safety constraint while violating the true safety
criterion.

\item We characterize how safety-filter contamination is amplified by reward maximization. We derive finite-$N$ bounds for \cbon{} in terms of the joint reward tails of safe and unsafe proxy-feasible outputs, and give conditions under which safety hacking becomes asymptotically certain even as unsafe false acceptances and average proxy errors vanish.

\item We derive $N$-independent safety-hacking bounds under coverage control and study \cPes{} as a concrete instance.
These bounds limit amplification but cannot repair contamination introduced by an imperfect safety constraint.

\item Controlled and language-model experiments measure both stages of the mechanism.
An exact finite-$N$ decomposition separates exposure to unsafe-but-feasible responses from the probability that they outrank every safe candidate.
The latter increases with scale and drives the increase in safety hacking.
Reward-model ablations isolate the role of reward ranking in amplification, and we observe the same decomposition pattern with an alternative safety filter.

\end{enumerate}

\section{Related Work}
\label{sec:related_work}

\textbf{Reward overoptimization.}
Optimizing an imperfect proxy can produce unintended behavior, a concern studied through reward hacking or Goodhart's law \citep{amodei2016concrete,manheim2018categorizing}.
In language-model alignment, learned reward models are used both during training \citep{ouyang2022training,bai2022constitutional} and to rank responses at inference time \citep{stiennon2020learning}.
Prior work shows that excessive optimization of an imperfect reward proxy, through reinforcement learning or inference-time scaling, can eventually reduce true reward \citep{gao2023scaling,khalaf2026inference}.
Our setting includes misspecification in both the reward and the safety constraint (i.e., optimization domain).
A learned safety proxy defines a feasible set that can contain unsafe false acceptances,
which reward maximization may then amplify through upper-tail selection.
We analyze the interaction between these two sources of error.

\textbf{Safety-constrained inference-time alignment.}
\citet{chittepu2026safe} formulate inference-time alignment with learned reward and safety cost models, using a calibrated Lagrangian reward for sequence-level \bon{} under an expected-cost constraint.
While \citet{chittepu2026safe} optimize against the learned cost signal, we instead study a misspecified hard safety filter.
In our settings, false acceptances contaminate the feasible set, and subsequent reward maximization can amplify them.

\textbf{Inference-time pessimism and coverage control.}
\citet{huang2025best} study coverage and scaling in inference-time alignment
and propose pessimistic alternatives to na\"ive \bon{}.
Also, regularized \bon{} has been studied under a minimum-Bayes-risk objective \citep{jinnai2025regularized}.
We condition the reference distribution on proxy feasibility and apply coverage control to the resulting distribution.
The bound limits downstream amplification of safety-filter errors but leaves any contamination already present in the
conditioned reference.
The distinction from prior work lies in separating contamination from amplification. 
Coverage control limits amplification after conditioning on proxy feasibility.

\textbf{Learned safety filters and evaluator errors.}
LLM systems often use learned filters, moderation models, or guardrails to classify prompts and responses.
Llama Guard casts safeguarding as safety-risk classification~\citep{inan2023llama}, while ShieldGemma provides open models for detecting risks in model inputs and outputs \citep{zeng2024shieldgemma}.
These systems make safety constraints operational but remain imperfect proxies for the true criterion \citep{zheng2023judging}.
We study how errors in the two proxies interact: unsafe false acceptances determine which responses enter the feasible set, and reward-proxy errors determine which response is selected.

\textbf{Jailbreaks and red teaming.}
HarmBench \citep{mazeika2024harmbench}, JailbreakBench \citep{chao2024jailbreakbench}, and StrongREJECT \citep{souly2024strongreject} provide benchmarks for harmful behavior, jailbreak, and refusal robustness.
\bon{} jailbreaking shows that repeated randomized attempts can substantially increase attack success rates \citep{hughes2024bestof}, demonstrating that search can expose rare safety failures.
We instead hold the prompt fixed and search over sampled model outputs.
The filter may admit unsafe outputs, and reranking can make them increasingly
likely to be selected.

\section{Problem Statement}
\label{sec:problem_statement}

We consider safety-constrained reward maximization at inference time.  
For clarity, we write the theory for a fixed prompt $x\in\Xcal$ whenever possible; hence,
all probabilities and expectations are conditional on this prompt.  
We take $\Ycal$ to be finite or countable, and the measurable-space extension is
obtained by replacing sums with integrals.
We suppose that there are an unknown reward function
$\rstar:\Xcal\times\Ycal\to\mbR_+$ and an unknown binary safety function
$\gstar:\Xcal\times\Ycal\to\{0,1\}$.  
For a prompt $x$, an output $y$ is truly
safe if $\gstar(x,y)=1$.
Define the following truly safe and unsafe sets of outputs:
\begin{align}
    \Scal_+^\star(x)
    &\coloneqq \{y\in\Ycal: \gstar(x,y)=1\},
    &
    \Scal_-^\star(x)
    &\coloneqq \Ycal\setminus\Scal_+^\star(x).
\end{align}
Thus, the ideal constrained inference problem is
\begin{equation}
    \label{eq:safety_constrained_problem}
    \max_{y\in\Ycal}\ \rstar(x,y)
    \qquad \text{subject to}\qquad
    \gstar(x,y)=1 .
\end{equation}
The agent cannot evaluate $\rstar$ or $\gstar$ directly, but instead has a
reference policy $\piref(\cdot\mid x)$, a learned reward proxy
$\rhat:\Xcal\times\Ycal\to\mbR$, and a learned safety proxy
$\ghat:\Xcal\times\Ycal\to[0,1]$.  
For a safety threshold $b\in(0,1]$, define the proxy-feasible set
\begin{align}
    \safehat(x,b)
    \coloneqq \{y\in\Ycal:\ghat(x,y)\ge b\}.
\end{align}
The proxy-feasible set decomposes into true positives and false positives:
\begin{align}
    A(x,b) &\coloneqq \Scal_+^\star(x)\cap\safehat(x,b),
    &
    B(x,b) &\coloneqq \Scal_-^\star(x)\cap\safehat(x,b).
\end{align}
Both sets contain outputs accepted by the safety proxy: $A(x,b)$ contains the truly safe ones, whereas $B(x,b)$ contains the unsafe ones.

\begin{definition}[Safety hacking]
\label{def:safety-hacking}
Fix $x\in\Xcal$ and $b\in(0,1]$.  
If an inference-time procedure outputs
$\widehat{y}\in\safehat(x,b)$, we say that safety hacking occurs when
\begin{align}
    \widehat{y}\in B(x,b).
\end{align}
Equivalently, the output passes the learned safety proxy but violates the true safety constraint.
Note that if the procedure abstains (e.g., when $\safehat(x,b) = \emptyset$), this is not considered safety hacking.
\end{definition}

The following joint upper-tail quantities are the central objects in the
CBo$N$ analysis.  For $\lozenge\in\{A,B\}$ and $t\in\mbR$, define the
learned-score tail
\begin{align}
    \widehat{\Psi}_{\lozenge}(t;x,b)
    \coloneqq
    \mbP_{y\sim\piref(\cdot\mid x)}
    \left[y\in \lozenge(x,b),\ \rhat(x,y)>t\right].
    \label{eq:joint-score-tail}
\end{align}
These tails combine 1)~the probability that the reference policy $\piref$ reaches a class of outputs $\lozenge$ and 2)~the upper-tail behavior of the learned reward score
inside that class.

Although we use a contextual bandit setting for simplicity, the formulation also covers open-loop trajectory selection. 
Let $\tau$ denote a trajectory.
Set $y=\tau$ and let $\piref(\tau\mid x)$ be the trajectory distribution
induced by a fixed reference policy and the environment dynamics, with reward
and safety defined at the trajectory level. The results require $N$ i.i.d.\
trajectories from this distribution and therefore exclude beam search, Monte
Carlo tree search, and other non-i.i.d.\ or stepwise procedures.

\section{Safety-Filter Contamination and Reward-Tail Amplification}
\label{sec:naive-constrained-bon-fails}

A natural baseline to solve \eqref{eq:safety_constrained_problem} is \cbon{}.
Given $N$ i.i.d. candidates $y_1,\ldots,y_N\sim\piref(\cdot\mid x)$, define a
proxy-feasible index set $\Ical_N(x,b)\coloneqq\{i\in[N]:y_i\in\safehat(x,b)\}$.
If $\Ical_N(x,b)\neq\varnothing$, CBo$N$ returns
\begin{equation}
    \widehat{y}_N^{\mathrm{BoN}}(x)
    \in
    \argmax_{i\in\Ical_N(x,b)} \rhat(x,y_i)
    \label{eq:naive_constrained_bon}
\end{equation}
with arbitrary tie-breaking.
In the case of $\Ical_N(x,b)=\varnothing$, the procedure
simply abstains.

The failure of \cbon{} has two stages. First, false positives from the safety
proxy contaminate the proxy-feasible set with outputs in $B(x,b)$. Second,
maximizing $\rhat$ over that set can amplify the contamination by searching the
extreme upper tail of the learned reward score. Increasing $N$ therefore does
more than increase exposure to unsafe-but-feasible outputs: it also intensifies
their reward-based competition with safe proxy-feasible outputs. The following
theorem identifies the joint score tails in \eqref{eq:joint-score-tail} as the
finite-$N$ quantities governing this competition.

\begin{theorem}[Finite-$N$ safety-hacking probability bounds]
\label{thm:hacking-finiteN}
Fix a prompt $x\in\Xcal$ and a threshold $b\in(0,1]$.  Let $y_1,\ldots,y_N\overset{\mathrm{iid}}{\sim}\piref(\cdot\mid x)$, and let
$\widehat{y}_N^{\mathrm{BoN}}(x)$ be defined by \eqref{eq:naive_constrained_bon}.
For any $t,s\in\mbR$, define
\begin{align}
    \widehat{L}_N(t;x,b)
    &\coloneqq
    \bigl(1-\widehat{\Psi}_A(t;x,b)\bigr)^N
    -
    \bigl(1-\widehat{\Psi}_A(t;x,b)-\widehat{\Psi}_B(t;x,b)\bigr)^N,
    \label{eq:finite-lower-bound}\\
    \widehat{U}_N(s;x,b)
    &\coloneqq
    1-
    \bigl(1-\widehat{\Psi}_B(s;x,b)\bigr)^N
    +
    \bigl(1-\widehat{\Psi}_B(s;x,b)-\widehat{\Psi}_A(s;x,b)\bigr)^N.
    \label{eq:finite-upper-bound}
\end{align}
Then, for every $t,s\in\mbR$,
\begin{align}
    \widehat{L}_N(t;x,b)
    \le
    \mbP\left[\widehat{y}_N^{\mathrm{BoN}}(x)\in B(x,b)\right]
    \le
    \widehat{U}_N(s;x,b).
\end{align}
Taking the supremum over $t$ and the infimum over $s$ gives the tightest bounds in this family.
\end{theorem}
The lower bound is the probability that no sample from $A(x,b)$ has learned
reward above $t$ and at least one sample from $B(x,b)$ does. 
CBo$N$ selects from $B(x,b)$ on this event. The upper bound follows because CBo$N$ cannot select from $B(x,b)$ when some sample from $A(x,b)$ scores above $s$ and none from $B(x,b)$ does. 
\Cref{cor:finiteN-operational} states the corresponding
guarantees in terms of the effective tail masses $N\widehat{\Psi}_A$ and $N\widehat{\Psi}_B$.
Let $\bar{t} \coloneqq \operatornamewithlimits{ess\,sup}_{y\in A(x,b)}\widehat r(x,y)$, so that $\widehat{\Psi}_A(\bar t;x,b)=0$. 
Theorem~\ref{thm:hacking-finiteN} then gives
$\mbP[\widehat{y}_N^{\mathrm{BoN}}(x)\in B(x,b)]
\ge 1-(1-\widehat{\Psi}_B(\bar{t};x,b))^N$.
If $\widehat{\Psi}_B(\bar t;x,b)=0.01$, the lower bound in
\Cref{thm:hacking-finiteN} is approximately $0.63$ at $N=100$ and $0.99$ at
$N=500$.

Contamination determines the mass of $B(x,b)$. The relative joint tails
determine whether CBo$N$ selects from it. The next result gives the following
tail-separation condition under which the safety-hacking probability converges
to one.
\begin{theorem}[Asymptotic safety hacking]
\label{thm:hacking-asymptotic}
Fix $x\in\Xcal$ and $b\in(0,1]$.  If there exists a sequence
$t_N\in\mbR$ such that
$N\widehat{\Psi}_A(t_N;x,b) \to 0$,
and
$N\widehat{\Psi}_B(t_N;x,b)\to\infty$,
then
\begin{align}
    \lim_{N\to\infty}
    \mbP\left[\widehat{y}_N^{\mathrm{BoN}}(x)\in B(x,b)\right]
    =1.
    \label{eq:safety_hacking_1}
\end{align}
\end{theorem}
In the boundary case discussed above,
$\widehat{\Psi}_A(\bar t;x,b)=0$ and
$\widehat{\Psi}_B(\bar t;x,b)>0$.
Setting $t_N=\bar t$, the theorem shows that the safety-hacking probability
converges to one.
More generally, even an arbitrarily small contaminated region can dominate
\bon{} search when the reward tails separate.
The fixed-prompt result extends directly to prompt distributions: if the tail-separation condition holds on a set of prompts with probability mass at least $\rho$, the aggregate hacking probability has liminf at least $\rho$; see \Cref{cor:prompt-distribution-amplification}.

\begin{corollary}[Vanishing proxy error still allows amplification]
\label{cor:small-rmse-amplification}
Fix $x$ and $b\in(0,1]$.  For every $\varepsilon\in(0,1)$ and $\xi>0$,
there exists a two-output instance whose unsafe-but-feasible reference mass is
$\varepsilon$, whose safety- and reward-proxy RMSEs are respectively
$b\sqrt{\varepsilon}$ and $\xi\sqrt{\varepsilon}$, and for which
\begin{align}
    \mbP \left[\widehat y_N^{\mathrm{BoN}}(x)\in B(x,b)\right]
    =1-(1-\varepsilon)^N
    \ge 1-e^{-N\varepsilon}.
\end{align}
Consequently, any
$N\ge \varepsilon^{-1}\log(1/\delta)$ yields safety hacking with probability
at least $1-\delta$.  Thus, even as the false-positive mass and both average
proxy errors vanish, a budget $N=\Theta(1/\varepsilon)$ can amplify the
residual error to high probability.
\end{corollary}

When the true-reward range over proxy-feasible outputs is bounded, tail
separation also follows if reward overestimation has a heavier Gaussian upper tail on unsafe-but-feasible outputs than on safe-feasible outputs. The relevant threshold scales as $\sqrt{\log N}$; see
\Cref{prop:gaussian-special-case}.

\section{Coverage Control Limits Amplification, Not Contamination}
\label{sec:pessimism-constrained}

The preceding analysis separates two sources of risk. \emph{Contamination} is
the unsafe mass admitted by the safety proxy; \emph{amplification} is the
additional concentration on that mass induced by downstream reward
optimization. 
This distinction suggests a general mitigation principle:
bounded deviation from the proxy-feasible reference distribution limits
amplification as $N$ grows, although it cannot remove the underlying
contamination.

Let
$q(x,b)\coloneqq\mbP_{y\sim\piref(\cdot\mid x)}[y\in\safehat(x,b)]$
denote the reference probability that a sampled output passes the learned
safety filter. 
Throughout this section, we assume $q(x,b)>0$. 
Conditioning the reference policy on this event gives the proxy-feasible reference distribution
\begin{align}
    \piref^\sharp(y\mid x)
    \coloneqq
    \frac{\piref(y\mid x) \cdot \ind\{y\in\safehat(x,b)\}}{q(x,b)}.
    \label{eq:def_pi_sharp}
\end{align}
This distribution is the baseline against which we measure how
strongly an inference procedure concentrates on feasible outputs.
We write $\pi(\cdot\mid x)\ll\piref^\sharp(\cdot\mid x)$ when
$\pi(y\mid x)>0$ implies $\piref^\sharp(y\mid x)>0$ for every $y\in\Ycal$.
For any such policy, define the following coverage coefficient by
\begin{align}
    C_\pi^\sharp(x)
    \coloneqq
    \sum_{y\in\Ycal}\frac{\pi(y\mid x)^2}{\piref^\sharp(y\mid x)}
    =1+\chi^2\left(
        \pi(\cdot\mid x)\,\middle\|\,\piref^\sharp(\cdot\mid x)
    \right).
    \label{eq:def-feasible-coverage}
\end{align}
Small $C_\pi^\sharp(x)$ means that the final policy retains broad coverage of
the proxy-feasible reference distribution, whereas a large value indicates
strong concentration. 

\begin{theorem}[Coverage-controlled contamination amplification]
\label{thm:coverage-controlled-safety}
Define the residual unsafe mass under $\piref^\sharp$ as
$\kappa(x,b) \coloneqq \mbP_{y\sim\piref^\sharp(\cdot\mid x)}[y\in B(x,b)]$.
Any proxy-feasible policy
$\pi(\cdot\mid x)\ll\piref^\sharp(\cdot\mid x)$ satisfies
\begin{align}
    \left|
        \mbP_{y\sim\pi(\cdot\mid x)}[y\in B(x,b)]
        -
        \kappa(x,b)
    \right|
    \le
    \sqrt{
        \bigl(C_\pi^\sharp(x)-1\bigr)
        \kappa(x,b)\bigl(1-\kappa(x,b)\bigr)
    }.
    \label{eq:contamination-amplification}
\end{align}
Define $\bar\varepsilon_g(x)\coloneqq\sqrt{
\mbE_{y\sim\piref(\cdot\mid x)}
\left[(\ghat(x,y)-\gstar(x,y))^2\right]}$. Then,
\begin{align}
    \kappa(x,b)
    \le
    \min \left\{1, \frac{\bar\varepsilon_g(x)^2}{b^2 q(x,b)} \right\}.
    \label{eq:coverage-controlled-safety-bound}
\end{align}
\end{theorem}

The theorem applies to any inference procedure with bounded proxy-feasible
coverage, regardless of how its final policy is constructed.
Thus coverage control limits how much downstream selection can amplify the
baseline contamination of the proxy-feasible set, but it cannot remove that
contamination itself. In particular, $C_\pi^\sharp(x)=1$ implies
$\pi=\piref^\sharp$ and the safety-hacking probability equals $\kappa(x,b)$.

\subsection{A Regularized Instantiation}
\label{subsec:regularized-instantiation}

To prevent arbitrarily large learned reward scores from inducing extreme
concentration, fix clipping levels
$-\infty<\widehat R_{\min}<\widehat R_{\max}<\infty$ and set $\widehat R_{\mathrm{span}} \coloneqq \widehat R_{\max}-\widehat R_{\min}$.
Then, define
\begin{align}
    \widetilde r(x,y)
    \coloneqq
    \operatorname{clip}\left(
        \rhat(x,y),
        \widehat R_{\min},
        \widehat R_{\max}
    \right).
    \label{eq:clipped-reward-score}
\end{align}
We use $\widetilde r$ only for the
coverage-controlled policy below; the CBo$N$ analysis in
\Cref{sec:naive-constrained-bon-fails} continues to use the original learned
score $\rhat$.

For $\beta>0$, consider the regularized proxy-feasible policy
\begin{equation}
\label{eq:pessimistic-policy-optimization}
\hat\pi
\in
\argmax_{\pi\ll\piref^\sharp}
\left\{
    \mbE_{\pi}[\widetilde r(x,y)]
    -\frac{\beta}{2}\bigl(C_\pi^\sharp(x)-1\bigr)
\right\}.
\end{equation}

\begin{proposition}[Regularized reweighting]
\label{prop:pessimistic-policy-closed-form}
There is a unique
$\lambda(x)\in[\widehat R_{\min}-\beta,\widehat R_{\max})$ satisfying
$\mbE_{y \sim \piref^\sharp(\cdot \mid x)}
\left[\relu\left(\frac{\widetilde r(x,y)-\lambda(x)}{\beta}\right)\right]=1$,
and the unique optimizer of \eqref{eq:pessimistic-policy-optimization} is
\begin{align}
    \hat\pi(y\mid x)
    =
    \piref^\sharp(y\mid x)
    \relu\left(
        \frac{\widetilde r(x,y)-\lambda(x)}{\beta}
    \right).
    \label{eq:pessimistic-constrained-policy}
\end{align}
Moreover, its concentration relative to the proxy-feasible reference is
bounded as
\begin{align}
    C_{\hat\pi}^\sharp(x)
    \le
    1+\widehat R_{\mathrm{span}}/\beta.
    \label{eq:coverage-bound-pessimistic-policy}
\end{align}
\end{proposition}
Hereinafter, we call the resulting algorithm constrained pessimistic sampling (\cPes).
A finite-sample implementation replaces the expectations above by their
empirical analogues over the proxy-feasible candidates and samples according
to the resulting reweighting. Full pseudocode and consistency are given in
\Cref{alg:cPes,prop:lambda-consistency}.
Let $m_N=|\Ical_N(x,b)|$ be the number of proxy-feasible candidates, and let
$\widehat y_{\mathrm{cPes},N}(x)$ denote the output of finite-sample \cPes{}.

\begin{corollary}[Safety of \cPes]
\label{cor:cPes-safety}
Conditional on $m_N>0$, the sample implementation satisfies
\begin{align}
    \mbP[
        \widehat y_{\mathrm{cPes},N}(x)\in B(x,b)
        \mid m_N>0
    ]
    \le
    \min\left\{
        1,\,
        \frac{\bar\varepsilon_g(x)^2}{b^2q(x,b)}
        +
        \frac{\bar\varepsilon_g(x)}{b}
        \sqrt{
            \frac{\widehat R_{\mathrm{span}}}
                 {\beta q(x,b)}
        }
    \right\}.
    \label{eq:safety-upperbound-main}
\end{align}
The same bounds hold for the population policy
\eqref{eq:pessimistic-constrained-policy}. With an abstaining fallback, the
sample bounds hold unconditionally for every $N$.
\end{corollary}

\section{Experiments}
\label{sec:experiments}

\subsection{Toy Problem}
\label{subsec:toy_experiment}

We illustrate tail separation with a three-class toy problem.
Each candidate belongs to a latent class $A$, $B$, or $C$
with probabilities $(0.39,0.01,0.60)$. Class $A$ is truly safe and
proxy-feasible, class $B$ is unsafe but proxy-feasible, and class $C$ is
correctly rejected by the safety proxy. We set
\[
    g^\star=(1,0,0),
    \qquad
    \widehat g=(0.95,0.90,0.05),
    \qquad
    b=0.8,
\]
and use true rewards $r^\star=(0.8,0.2,0.5)$. The learned reward is
\[
    \hat r
    =
    r^\star + \sigma_k Z,
    \qquad
    Z\sim\mathcal N(0,1),
\]
with $(\sigma_A,\sigma_B,\sigma_C)=(0.2,1.0,0.2)$. Thus the unsafe
false-positive class $B$ is rare, but has a substantially heavier upper tail
of reward-proxy overestimation than the safe class $A$. We compare random
feasible selection, CBo$N$, and \cPes{} over increasing candidate budgets.
Random feasible selection chooses uniformly among the candidates accepted by
the safety proxy, without using the learned reward.
\cbon{} uses the unclipped learned score $\hat r$, whereas cPes applies the clipping with $[\widehat R_{\min},\widehat R_{\max}]=[0,4]$.

\begin{figure*}[t]
    \centering
    \begin{subfigure}[t]{0.32\textwidth}
        \centering
        \includegraphics[width=\linewidth]{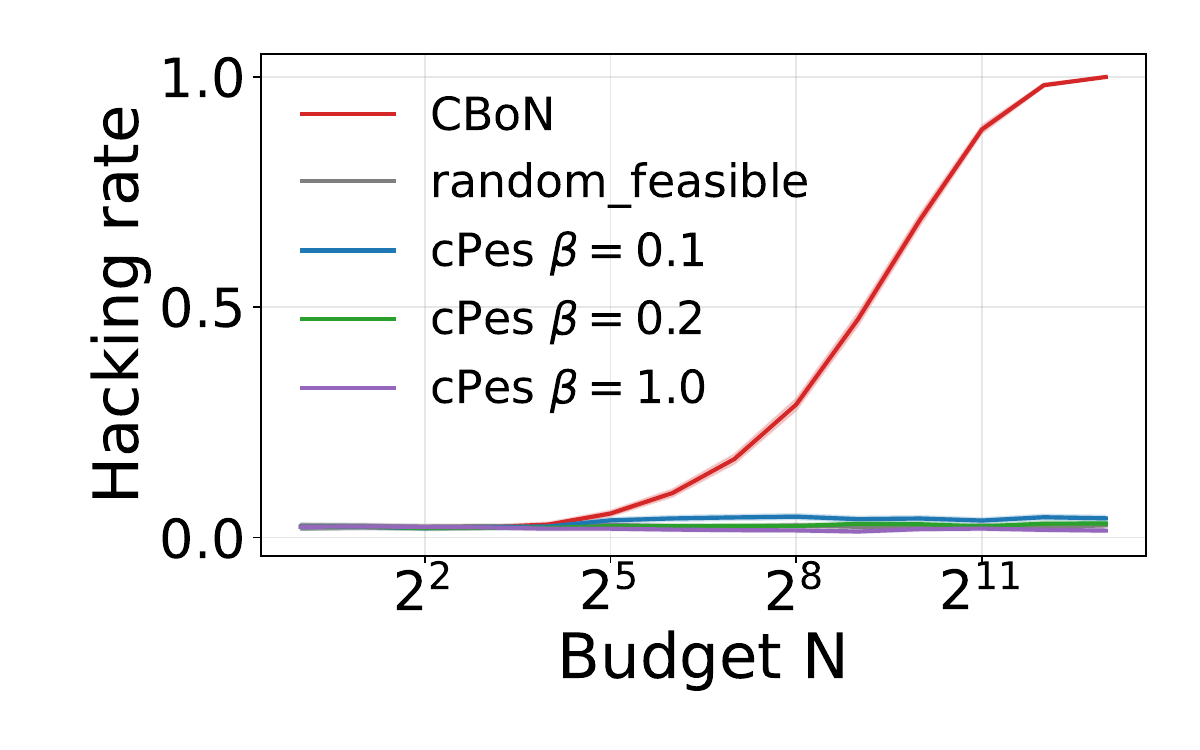}
        \caption{Conditional safety-hacking rate.}
    \end{subfigure}
    \hfill
    \begin{subfigure}[t]{0.32\textwidth}
        \centering
        \includegraphics[width=\linewidth]{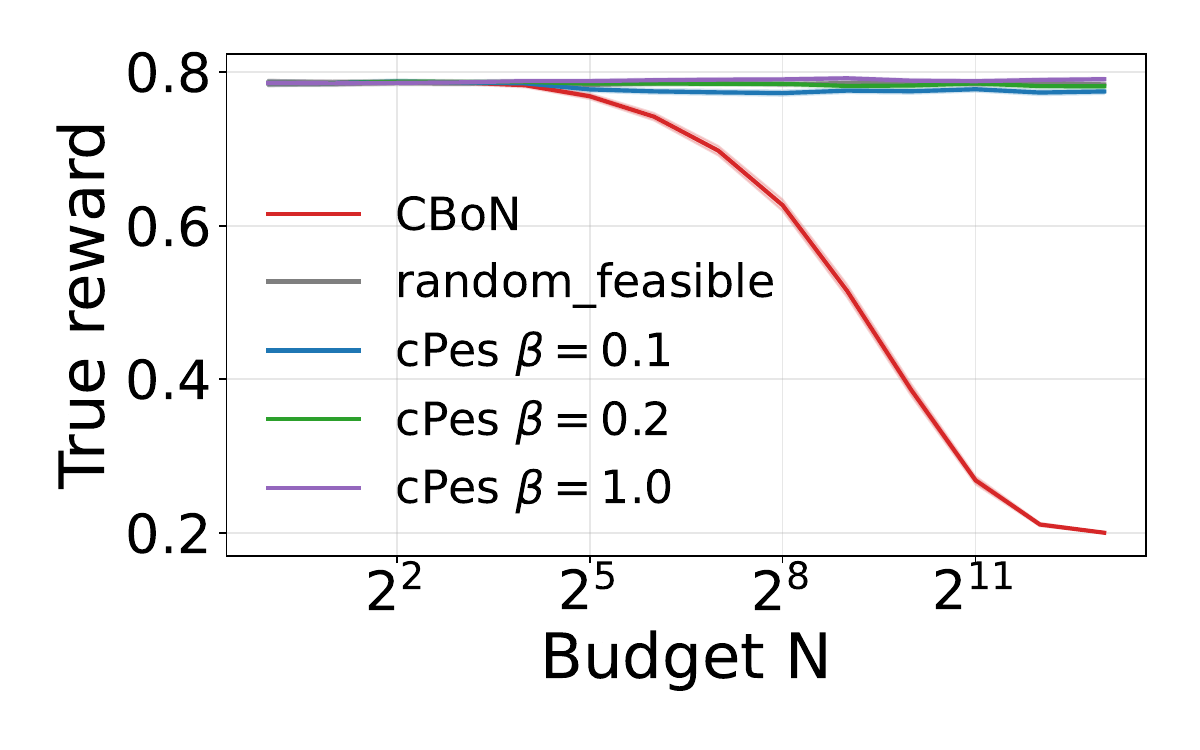}
        \caption{Conditional mean true reward.}
    \end{subfigure}
    \hfill
    \begin{subfigure}[t]{0.32\textwidth}
        \centering
        \includegraphics[width=\linewidth]{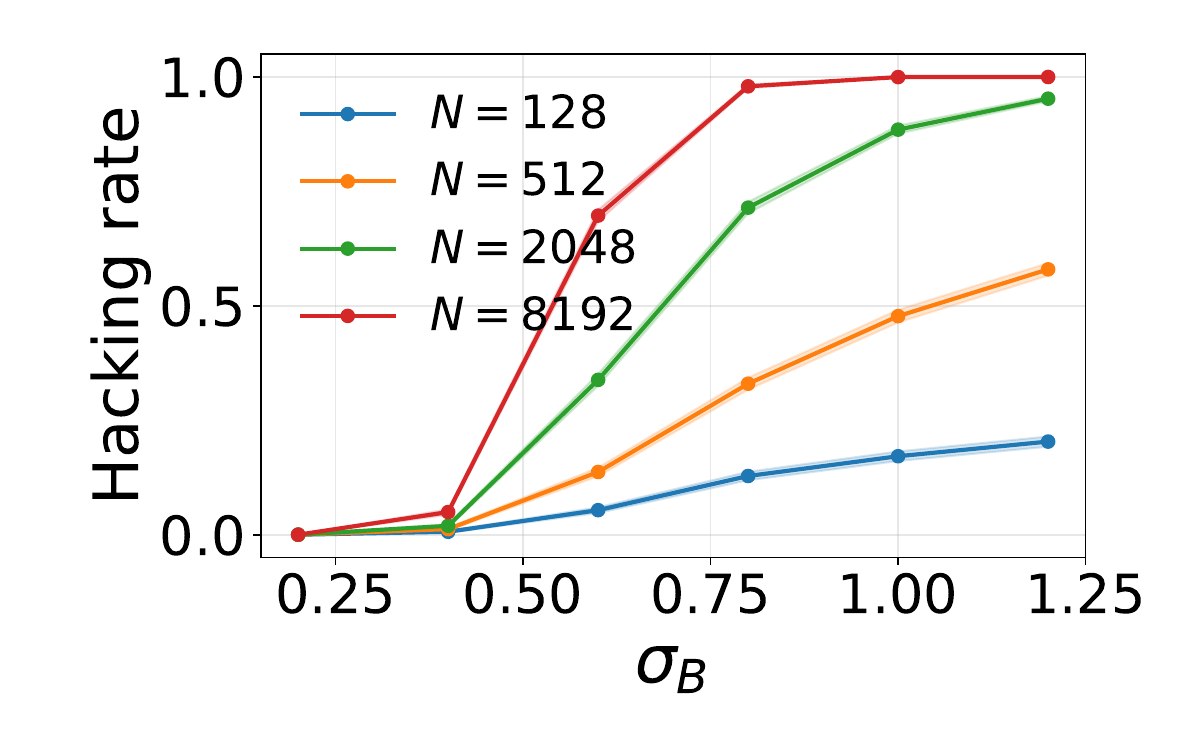}
        \caption{False-positive tail scale.}
        \label{fig:toy-sigma-b}
    \end{subfigure}
    \caption{
        Toy results. As the search budget grows, CBo$N$ increasingly selects
        the rare unsafe-but-feasible class $B$, whose reward-proxy upper tail
        dominates that of the safe class $A$. Random feasible selection does
        not exhibit this amplification, while \cPes{} limits the resulting
        concentration. Panel (c) varies the false-positive reward-noise scale
        $\sigma_B$.
    }
    \label{fig:toy-main}
\end{figure*}

\Cref{fig:toy-main} shows the predicted amplification. 
As $N$ grows, CBo$N$ shifts from selecting the safe class $A$ to the unsafe-but-feasible class $B$.
The conditional safety hacking rate rises from $0.023$ at $N=1$ to $0.9996$ at $N=8192$, while the mean true reward falls from about $0.786$ to $0.200$.
Random feasible selection remains near the baseline false-positive rate,
showing that exposure to class $B$ alone does not explain the failure.
\cPes{} substantially limits this amplification in the toy problem.
Controls varying the reward-error tails and removing false positives confirm
that the effect requires an unsafe proxy-feasible class with a sufficiently
advantageous learned-reward tail; see Appendix~\ref{app:toy-additional}.

\subsection{LLM Experiments}
\label{subsec:llm_experiment}

We next evaluate safety hacking in a more practical setting using LLMs.

\textbf{Experimental settings.}
Our evaluation uses 714 test and 179 validation prompts from
JailbreakBench~\citep{chao2024jailbreakbench},
HarmBench~\citep{mazeika2024harmbench}, and
AdvBench~\citep{chen2022should}, after removing duplicates. 
For each prompt,
we sample 256 responses from \texttt{Qwen/Qwen2.5-7B-Instruct}~\citep{qwen2.5} (temperature
1.0, top-$p$ 0.95, 512 new tokens), giving 228{,}608 candidates. 
We evaluate
all selection methods on the same candidates.

We filter candidates with \nolinkurl{meta-llama/Llama-Guard-3-8B}~\citep{dubey2024llama3herdmodels} and rank them
with \nolinkurl{PKU-Alignment/beaver-7b-v1.0-reward}~\citep{dai2024safe}. We report the primary safety
threshold $b=0.95$ and compare random feasible selection, \cbon{}, and \cPes{}
for $N\in\{1,2,4,8,16,32,64,128,256\}$. 
Random feasible selection controls for exposure to filter false positives without reward maximization; \cPes{} uses
calibration-fixed clipping levels and $\beta=1.0$. 
Point estimates are averaged over 100 random candidate-order permutations. 
For each prompt, we first average the metric over permutations and then compute standard errors across prompts; shaded bands show $\pm 1.96$ standard errors.

Because latent safety and reward are unobservable for open-ended responses, we
use the HarmBench classifier as the operational safety criterion and define a
safety-hacking event as selecting a response that passes Llama Guard but is
classified as unsafe by HarmBench. Reported safety-hacking rates condition on
non-abstention. Separately, we use \texttt{gpt-5-mini}~\citep{singh2025openai} to evaluate
safety-aware reward under the \texttt{safety\_aware\_v1} rubric. These
evaluators operationalize $g^\star$ and $r^\star$ for the experiments; neither
is treated as ground truth. Evaluator and scoring details are given in
Appendix~\ref{app:llm-evaluation-details}. 

\textbf{Safety-hacking amplification and finite-$N$ decomposition.}
For a fixed prompt, candidate ordering, and budget $N$, let $K_A$ and $K_B$ be
the numbers of proxy-feasible candidates labeled safe and unsafe by HarmBench.
When the corresponding class is nonempty, let $M_A$ and $M_B$ denote its largest proxy reward. 
Because the proxy scores have no ties, CBo$N$ selects an unsafe response in one of two cases: no safe feasible candidate is available,
or the best unsafe candidate outscores the best safe candidate. 
Conditional on non-abstention, this gives
\begin{align}
&\Pr\left[
    \widehat y_N^{\mathrm{BoN}}\in B(x,b)
    \mid K_A+K_B>0
\right]
\nonumber \\
&=
\Pr\left[
    K_B>0,\ K_A=0
    \mid K_A+K_B>0
\right] +
\Pr \left[
    K_A>0,\ K_B>0,\ M_B>M_A
    \mid K_A+K_B>0
\right].
\label{eq:finite-n-hacking-decomposition}
\end{align}
The first term measures unsafe-only exposure, where no safe feasible candidate
is available. 
The second measures reward-based competition when both classes are present, and is the empirical finite-$N$ counterpart of the joint-tail competition in \Cref{thm:hacking-finiteN}.

\begin{figure}[t]
    \centering
    \includegraphics[width=\linewidth]{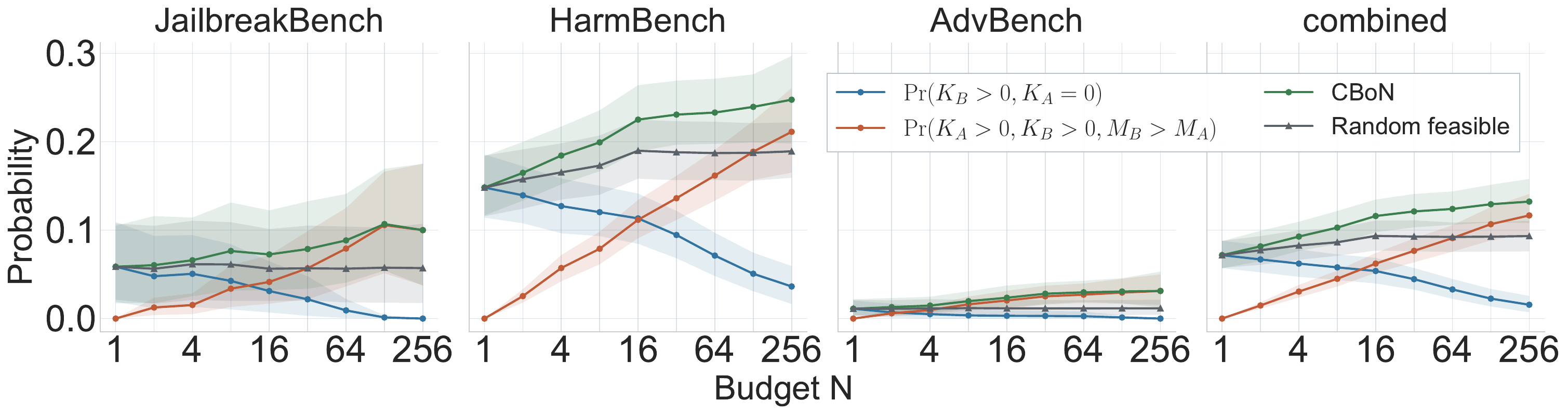}
    \caption{Finite-$N$ decomposition of conditional CBo$N$ safety hacking on
    adversarial prompts under the HarmBench operational safety criterion. The
    total (green) is the sum of unsafe-only exposure (blue) and competitive
    unsafe wins (orange); random feasible selection is shown in gray. The
    competitive term corresponds to the joint-tail mechanism in
    \Cref{thm:hacking-finiteN}. Points are averaged over $100$ candidate-order permutations; bands show $\pm 1.96$ standard errors across prompts after permutation averaging.}
    \label{fig:exposure-tail-decomposition}
\end{figure}

\begin{figure*}[t]
\centering
\begin{subfigure}[t]{0.32\textwidth}
\centering
\includegraphics[width=\linewidth]{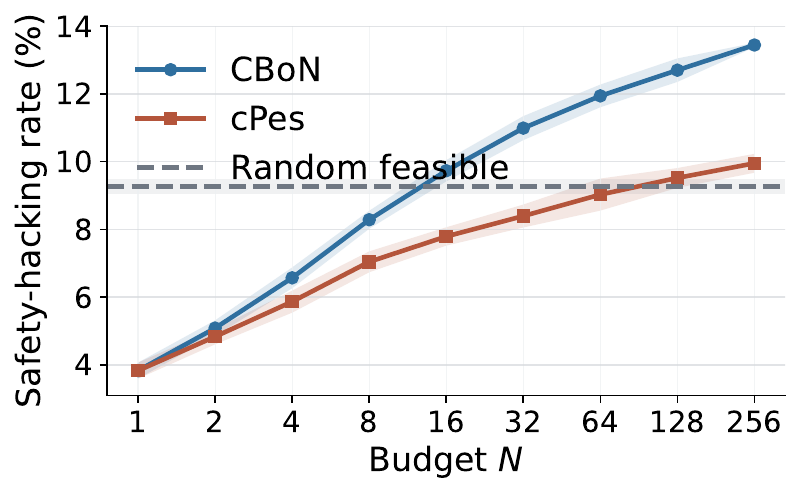}
\caption{Safety-hacking rate.}
\label{fig:llm-safety-hacking}
\end{subfigure}
\hfill
\begin{subfigure}[t]{0.32\textwidth}
\centering
\includegraphics[width=\linewidth]{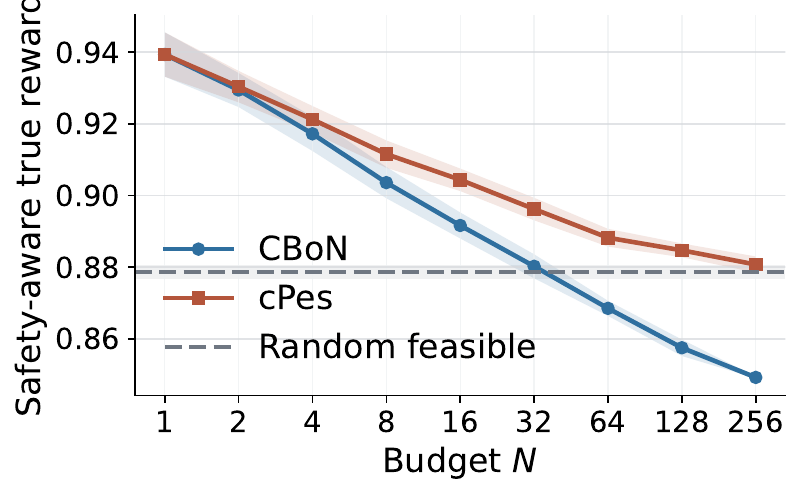}
\caption{Safety-aware true reward.}
\label{fig:llm-safety-aware-reward}
\end{subfigure}
\hfill
\begin{subfigure}[t]{0.32\textwidth}
\centering
\includegraphics[width=\linewidth]{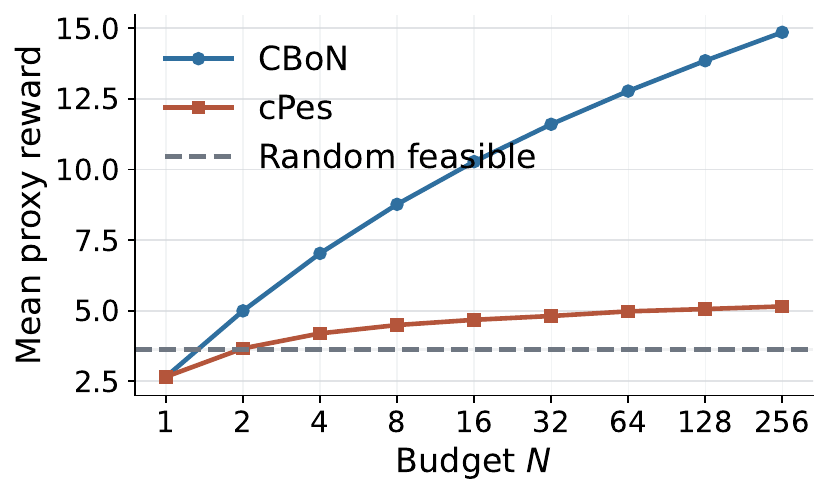}
\caption{Mean proxy reward.}
\label{fig:llm-proxy-reward}
\end{subfigure}
\caption{Safety and reward under constrained inference-time scaling on
adversarial prompts. HarmBench evaluates safety, and \texttt{gpt-5-mini}
evaluates safety-aware reward. As $N$ grows, CBo$N$ improves proxy reward but
worsens both safety measures, while \cPes{} limits these effects. Points are averaged over 100 candidate-order permutations; bands show $\pm 1.96$ standard errors across prompts after permutation averaging.}
\label{fig:llm-adversarial-main}
\end{figure*}

The decomposition in Figure~\ref{fig:exposure-tail-decomposition} attributes
the increase in safety hacking primarily to competitive selection. For the
HarmBench subset and the combined evaluation, the unsafe-only term decreases
with $N$, indicating that increased exposure to unsafe proxy-feasible responses
does not account for the observed scaling behavior. 
By contrast, the competitive-selection term increases: conditional on both classes being present, the unsafe class more frequently attains the larger maximum proxy reward. 
This trend is consistent with the finite-$N$ behavior implied by the relative joint tails in \Cref{thm:hacking-finiteN}.
Accordingly, the CBo$N$ hacking rate increasingly exceeds the random-feasible contamination baseline, whereas \cPes{} remains closer to the baseline (\Cref{fig:llm-safety-hacking}).
Repeating the analysis with a calibration-matched \texttt{google/shieldgemma-2b}~\citep{zeng2024shieldgemmagenerativeaicontent} filter gives
the same pattern, but a smaller overall increase in safety hacking: unsafe-only
exposure falls with $N$, while competitive unsafe wins become more common
(Appendix~\ref{app:shieldgemma-filter-robustness}).

\textbf{Proxy reward versus safety-aware reward.}
Larger budgets improve the objective optimized by CBo$N$: its proxy reward
increases well above the random-feasible reference
(\Cref{fig:llm-proxy-reward}). Yet its \texttt{gpt-5-mini}-judged safety-aware reward, which
penalizes unsafe and non-responsive outputs, deteriorates with $N$
(\Cref{fig:llm-safety-aware-reward}). \cPes{} produces smaller departures on
both measures. Thus aggressive proxy optimization exploits proxy errors rather
than improving the intended objective.

\begin{figure*}[t]
    \centering
    \begin{subfigure}[t]{0.42\textwidth}
        \centering
        \includegraphics[width=\linewidth]{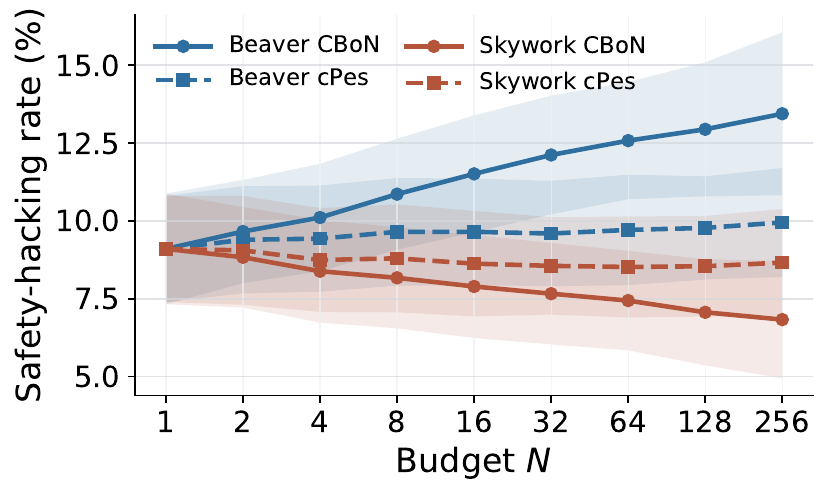}
        \caption{Safety-hacking rate.}
        \label{fig:skywork-hacking-scaling}
    \end{subfigure}
    \hfill
    \begin{subfigure}[t]{0.56\textwidth}
        \centering
        \includegraphics[width=\linewidth]{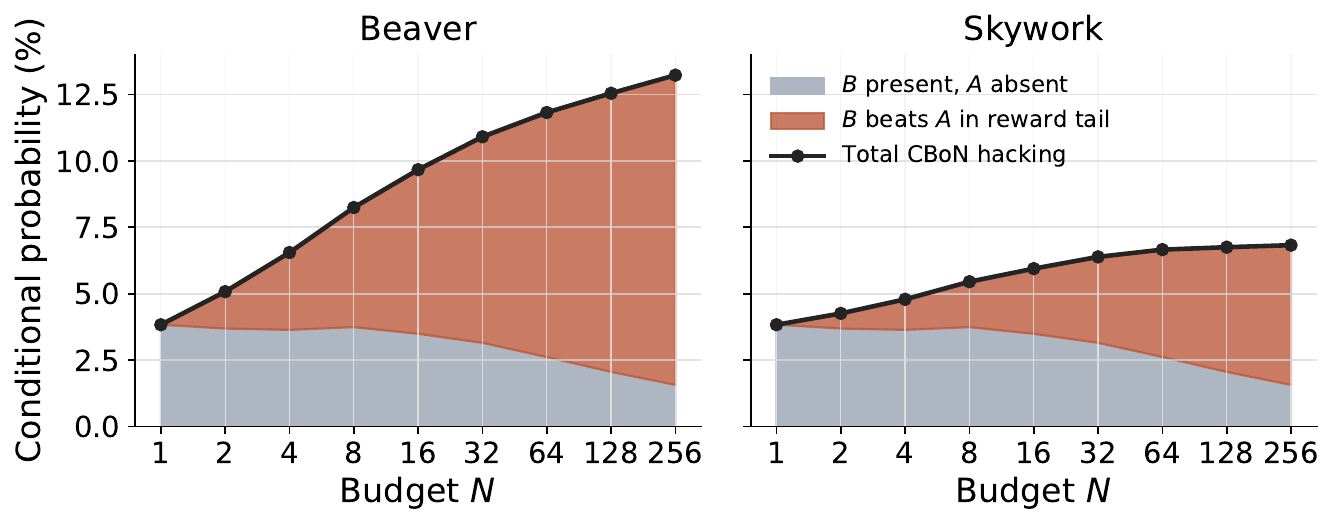}
        \caption{Finite-$N$ decomposition for CBo$N$.}
        \label{fig:skywork-finite-n-decomposition}
    \end{subfigure}
    \caption{Reward-proxy ablation with fixed candidate pools, safety-filter
    decisions, and HarmBench labels. Replacing Beaver with Skywork reverses the scaling trend in CBo$N$ safety hacking (left). The unsafe-only component is unchanged, whereas the competitive component decreases under Skywork (right), indicating that amplification depends on reward-based competition
    within the contaminated proxy-feasible set.}
    \label{fig:skywork-reward-ablation}
\end{figure*}

\textbf{Reward-proxy ablation.}
To isolate the effect of the reward proxy, we replace
\nolinkurl{PKU-Alignment/beaver-7b-v1.0-reward} with
\nolinkurl{Skywork/Skywork-Reward-V2-Llama-3.1-8B}~\citep{liu2025skywork}, while keeping the candidate
pools, safety filter, and HarmBench labels fixed. With Beaver, CBo$N$ safety
hacking increases from $9.1\%$ at $N=1$ to $13.4\%$ at $N=256$; with Skywork,
it decreases to $6.8\%$. At $N=256$, the unsafe-only term is $1.6\%$ under both
proxies, whereas the competitive unsafe-win term decreases from $11.7\%$ with
Beaver to $5.3\%$ with Skywork (\Cref{fig:skywork-reward-ablation}). The reversal
is therefore attributable to how the two reward proxies rank the same safe and
unsafe proxy-feasible candidates, rather than to filtering or exposure. The
corresponding conditional reward-tail survival curves are reported in
\Cref{fig:skywork-tail-survival}.

This ablation is the empirical counterpart of the finite-$N$ bounds in
\Cref{thm:hacking-finiteN,cor:finiteN-operational}, which depend on the joint
tails $\widehat{\Psi}_A$ and $\widehat{\Psi}_B$ rather than false-positive mass alone. 
A favorable safe tail can prevent amplification over the evaluated budgets without providing an asymptotic guarantee. 
At $N=256$, Skywork CBo$N$ has a lower safety-hacking rate than Skywork \cPes{} ($6.8\%$ versus $8.7\%$).
The guarantee for \cPes{} is therefore not pointwise dominance over CBo$N$, but an $N$-independent bound under adverse tail configurations (\Cref{cor:cPes-safety} and \Cref{thm:coverage-controlled-safety}).

\section{Discussion and Limitations}
\label{sec:discussion}

\textbf{Coverage control limits amplification, not contamination or
misranking.}
The proxy-feasible reference already assigns mass $\kappa(x,b)$ to unsafe
false positives. \cPes{} limits further concentration on this residual mass
but cannot identify it, and may still favor unsafe outputs with high learned
rewards. Its $N$-independent bound therefore implies neither absolute safety
nor pointwise dominance over \cbon{}, as illustrated by the Skywork ablation.

\textbf{Robust inference requires intervention at multiple points.}
Our analysis points to three complementary interventions: improving the safety
filter, limiting downstream concentration, and choosing reward proxies with
favorable safe-versus-unsafe tail behavior. The Skywork ablation shows that
reward-tail ordering can substantially change risk over the observed budget
range. These interventions address different parts of the failure: filtering
controls contamination, while reward-tail behavior and coverage control govern
selection within the resulting feasible set.

\textbf{Limitations.}
We analyze a fixed, non-adaptive pipeline in which independently sampled candidates are filtered once and then ranked. Our theory therefore does not cover adaptive search, self-refinement, tree search, or agentic planning,
whose candidate distributions can depend on previous proxy evaluations.
By repeatedly steering generation toward high proxy scores, such procedures may amplify residual safety-filter errors more strongly than i.i.d. scaling,
but our results do not establish or quantify this behavior. Extending the contamination--amplification analysis to such feedback-driven procedures is an important direction for future work.

\section{Conclusion}
\label{sec:conclusion}

We studied safety hacking in constrained inference-time scaling, where reward
maximization can amplify residual errors in a learned safety filter. The filter
may admit unsafe responses, and downstream selection may favor them more
strongly as inference compute grows. Our finite-$N$ analysis shows that this
behavior is governed by the joint reward tails of safe and unsafe
proxy-feasible responses, and our asymptotic result establishes that a heavier
unsafe tail can make safety hacking nearly certain even when average proxy
errors are small. Coverage control yields an $N$-independent bound, instantiated
by \cPes{}, but cannot remove unsafe responses already admitted by the filter.
Toy and language-model experiments provide finite-budget evidence for this
amplification mechanism, while the reward-proxy ablation shows that ranking
within the contaminated feasible set is central to the observed outcome. The
analysis therefore separates two requirements for safe inference-time scaling:
controlling entry into the feasible set and controlling downstream
concentration. Extending this characterization to adaptive and agentic search
remains an important direction for future work.

\clearpage

\bibliography{main}
\bibliographystyle{plainnat}

\appendix
\section{Appendix}
\label{app:additional-results}

\subsection{Operational Finite-$N$ Safety-hacking Criteria}
\label{app:additional-cbon-results}

\begin{corollary}
\label{cor:finiteN-operational}
Under the conditions of \Cref{thm:hacking-finiteN}, the following hold.
\begin{enumerate}
    \item If there exist $t\in\mbR$, $\alpha_A\in[0,1)$, and $\gamma_B > 0$ such that
    $N\widehat{\Psi}_A(t;x,b) \le \alpha_A$ and 
    $N\widehat{\Psi}_B(t;x,b) \ge \gamma_B$,
    then
    \begin{align}
        \mbP \left[\widehat{y}_N^{\mathrm{BoN}}(x)\in B(x,b)\right]
        \ge
        (1-\alpha_A)\bigl(1-e^{-\gamma_B}\bigr).
    \end{align}
    \item If there exist $s\in\mbR$, $\alpha_B\ge0$, and $\gamma_A>0$ such that
    $N\widehat{\Psi}_B(s;x,b) \le \alpha_B$,
    and $N\widehat{\Psi}_A(s;x,b) \ge \gamma_A$,
    then
    \begin{align}
        \mbP\left[\widehat{y}_N^{\mathrm{BoN}}(x)\in B(x,b)\right]
        \le
        \alpha_B+e^{-\gamma_A}.
    \end{align}
\end{enumerate}
\end{corollary}

\begin{proof}
Let $\psi_A = \widehat{\Psi}_A(t;x,b)$ and $\psi_B = \widehat{\Psi}_B(t;x,b)$.  
From \Cref{thm:hacking-finiteN},
\begin{align}
    \widehat{L}_N(t;x,b)
    &=(1-\psi_A)^N\left[1-\left(1-\frac{\psi_B}{1-\psi_A}\right)^N\right].
\end{align}
If $N \psi_A \le \alpha_A < 1$, then $(1- \psi_A)^N \ge 1-N \psi_A \ge 1-\alpha_A$.
Since
$\psi_B/(1- \psi_A) \ge \psi_B$ and $N \psi_B \ge\gamma_B$,
\begin{align}
    1-\left(1-\frac{\psi_B}{1- \psi_A}\right)^N
    \ge
    1-e^{-N \psi_B}
    \ge
    1-e^{-\gamma_B}.
\end{align}
This proves the first claim.  
For the second claim, let
$\psi_A=\widehat{\Psi}_A(s;x,b)$ and $\psi_B=\widehat{\Psi}_B(s;x,b)$.  
The upper bound gives
\begin{align}
    \widehat{U}_N(s;x,b)
    &\le
    \bigl[1-(1-\psi_B)^N\bigr]+(1- \psi_A)^N
    \le
    N\psi_B + e^{-N \psi_A}
    \le
    \alpha_B+e^{-\gamma_A}.
\end{align}
This completes the proof.
\end{proof}

\Cref{cor:finiteN-operational} gives an experimentally testable diagnostic: as $N$ increases, estimate whether the effective unsafe-but-feasible score-tail mass $N\widehat{\Psi}_B(\cdot)$ crosses a constant before the true-positive score-tail mass $N\widehat{\Psi}_A(\cdot)$ does. 
If so, CBo$N$ should display compute-amplified safety hacking.

\subsection{Prompt-distribution Amplification}
\begin{corollary}[Prompt-distribution amplification]
\label{cor:prompt-distribution-amplification}
Let $X\sim\mathcal D$ be a prompt drawn from a prompt distribution, and fix
$b\in(0,1]$.  For each prompt $x$, define the conditional CBo$N$ hacking
probability, where the probability is over the $N$ candidate samples drawn from
$\piref(\cdot\mid x)$, by
\begin{align}
    p_N(x)
    \coloneqq
    \mbP_{y_{1:N}\sim\piref(\cdot\mid x)}\left[
        \widehat y_N^{\mathrm{BoN}}(x)\in B(x,b)
    \right].
\end{align}
Suppose there exist a measurable set of prompts $\mathcal E\subseteq\Xcal$ and
$\rho\in[0,1]$ such that
\[
    \mbP_{X\sim\mathcal D}[X\in\mathcal E]\ge \rho,
\]
and, for every $x\in\mathcal E$, there exists a sequence
$t_N(x)\in\mbR$ satisfying
$N\widehat{\Psi}_A(t_N(x);x,b)\to 0$ and 
$N\widehat{\Psi}_B(t_N(x);x,b)\to\infty$.
Then the aggregate safety-hacking probability satisfies
\begin{align}
    \liminf_{N\to\infty}
    \mbP\left[
        \widehat y_N^{\mathrm{BoN}}(X)\in B(X,b)
    \right]
    \ge
    \rho,
\end{align}
where the probability is over $X\sim\mathcal D$ and the candidate samples drawn
conditionally from $\piref(\cdot\mid X)$.
\end{corollary}

\begin{proof}
By \Cref{thm:hacking-asymptotic}, for every $x\in\mathcal E$ we have
$p_N(x)\to1$.  Moreover, $0\le p_N(x)\le1$ for all $x$ and $N$.  Therefore
\begin{align}
    \mbP\left[
        \widehat y_N^{\mathrm{BoN}}(X)\in B(X,b)
    \right]
    =
    \mbE_{X\sim\mathcal D}[p_N(X)]
    \ge
    \mbE_{X\sim\mathcal D}[p_N(X)\ind\{X\in\mathcal E\}].
\end{align}
Fatou's lemma gives
\begin{align}
    \liminf_{N\to\infty}
    \mbE[p_N(X)\ind\{X\in\mathcal E\}]
    &\ge
    \mbE\left[\liminf_{N\to\infty}p_N(X)\ind\{X\in\mathcal E\}\right]
    \\
    &=
    \mbP_{X\sim\mathcal D}[X\in\mathcal E]
    \ge
    \rho.
\end{align}
This proves the claim.
\end{proof}

\Cref{cor:prompt-distribution-amplification} lifts the fixed-prompt tail
condition to distribution-level evaluations: if a positive-measure subset of
prompts is tail-amplifiable, then the average safety-hacking rate cannot vanish
as the search budget grows.
\subsection{Direct score dominance}
\begin{remark}[Direct score dominance]
\label{rem:direct-score-dominance}
If there exists a threshold $t\in\mbR$ such that
$\widehat\Psi_A(t;x,b)=0$ and $\widehat\Psi_B(t;x,b)>0$, then
\Cref{thm:hacking-asymptotic} applies with the constant sequence $t_N=t$, and
\begin{align}
    \mbP[\widehat y_N^{\mathrm{BoN}}(x)\in B(x,b)]\to1.
\end{align}
This holds, for example, when a positive-reference-mass subset of $B(x,b)$
receives learned scores above the essential upper endpoint on $A(x,b)$.
\end{remark}
\subsection{Gaussian error-tail separation}
\begin{definition}[Proxy-feasible reward range]
\label{def:reward_gap}
Fix $x\in\Xcal$ and $b\in(0,1]$ with $\safehat(x,b)\neq\varnothing$.  Define
\begin{align}
    \Delta(x,b)
    \coloneqq
    \sup_{y\in\safehat(x,b)}\rstar(x,y)
    -
    \inf_{y\in\safehat(x,b)}\rstar(x,y).
\end{align}
\end{definition}

\begin{assumption}[Local bounded reward range]
\label{ass:local_reward_range}
For the prompt $x \in \Xcal$ and threshold $b \in (0,1]$ under consideration,
$\safehat(x,b)\neq\varnothing$ and $\Delta(x,b)<\infty$.
\end{assumption}

\begin{proposition}[Gaussian error-tail separation]
\label{prop:gaussian-special-case}
We define the pointwise proxy errors
\begin{align}
    \epsr(x,y) &\coloneqq \rhat(x,y)-\rstar(x,y), \\
    \epsg(x,y) &\coloneqq \ghat(x,y)-\gstar(x,y).
\end{align}
Also, for $\lozenge \in \{A,B\}$, define
\begin{align}
    \eta_\lozenge(x,b)
    \coloneqq
    \mbP_{y \sim \piref(\cdot \mid x)}[y \in \lozenge(x,b)] .
\end{align}
Fix $x \in \mathcal X$ and $b \in (0,1]$, and suppose
\Cref{ass:local_reward_range} holds.  Assume that
$\eta_A(x,b)>0$ and 
$\eta_B(x,b)>0$.
Suppose further that there exist constants
$0 < \sigma_A(x) < \sigma_B(x)$
such that the conditional upper tails of the reward-proxy error satisfy, as
$u\to\infty$,
\begin{align}
    \log
    \mbP\left[
        \epsr(x,y)>u
        \mid
        y\in A(x,b)
    \right]
    &=
    -\frac{u^2}{2\sigma_A^2(x)}
    + o(u^2),
    \\
    \log
    \mbP\left[
        \epsr(x,y)>u
        \mid
        y\in B(x,b)
    \right]
    &=
    -\frac{u^2}{2\sigma_B^2(x)}
    + o(u^2).
\end{align}
Let
\begin{align}
    R_A^\star(x,b)
    \coloneqq
    \sup_{y\in A(x,b)}\rstar(x,y).
\end{align}
Then, for any constant
$c \in \left(\sqrt{2}\sigma_A(x),\sqrt{2}\sigma_B(x)\right)$, the sequence
\begin{align}
    \tau_N
    \coloneqq
    R_A^\star(x,b)+c\sqrt{\log N}
\end{align}
satisfies the score-tail separation condition in
\Cref{thm:hacking-asymptotic}; that is,
\begin{align}
    N\widehat{\Psi}_A(\tau_N;x,b)\to0,
    \qquad
    N\widehat{\Psi}_B(\tau_N;x,b)\to\infty.
\end{align}
Consequently,
\begin{align}
    \lim_{N\to\infty}
    \mbP\left[
        \widehat{y}_N^{\mathrm{BoN}}(x)\in B(x,b)
    \right]
    =1.
\end{align}
The fixed shift $\Delta(x,b)$ only enters when converting reward-error tails into
learned-score tails, and it does not change the Gaussian large-deviation
exponent.
\end{proposition}

\begin{proof}

For $\lozenge\in\{A,B\}$ and $t\in\mbR$, define the joint reward-error tail
\begin{align}
    \Psi_{\lozenge}(t;x,b)
    \coloneqq
    \mbP_{y\sim\piref(\cdot\mid x)}
    \left[y\in \lozenge(x,b),\ \epsr(x,y)>t\right].
    \label{eq:joint-error-tail}
\end{align}
These tails combine the reference probability of a class with the upper-tail
behavior of reward-proxy overestimation inside that class.

For $\lozenge\in\{A,B\}$, we can write
\[
    \Psi_\lozenge(u;x,b)
    =
    \eta_\lozenge(x,b)
    \mbP\left[
        \epsr(x,y)>u
        \mid
        y\in \lozenge(x,b)
    \right].
\]
Let
\[
    t_N=c\sqrt{\log N},
    \qquad
    c\in(\sqrt{2}\sigma_A(x),\sqrt{2}\sigma_B(x)),
\]
and set
\[
    \tau_N=R_A^\star(x,b)+t_N.
\]
Using the assumed tail exponent on $A(x,b)$,
\begin{align}
    \log\left(N\Psi_A(t_N;x,b)\right)
    &=
    \log N+\log\eta_A(x,b)
    -\frac{t_N^2}{2\sigma_A^2(x)}
    +o(t_N^2)
    \\
    &=
    \left(
        1-\frac{c^2}{2\sigma_A^2(x)}+o(1)
    \right)\log N
    +O(1).
\end{align}
Since $c>\sqrt{2}\sigma_A(x)$, the coefficient of $\log N$ is negative,
and therefore
\begin{align}
    N\Psi_A(t_N;x,b)\to 0.
\end{align}
Moreover, for $y\in A(x,b)$, $\rstar(x,y)\le R_A^\star(x,b)$.  Hence
\begin{align}
    \{y\in A(x,b),\ \rhat(x,y)>\tau_N\}
    \subseteq
    \{y\in A(x,b),\ \epsr(x,y)>t_N\},
\end{align}
and so
\begin{align}
    N\widehat{\Psi}_A(\tau_N;x,b)
    \le
    N\Psi_A(t_N;x,b)
    \to0.
\end{align}

For the $B$ tail, since $\Delta(x,b)<\infty$ by assumption,
\begin{align}
    (t_N+\Delta(x,b))^2
    =
    c^2\log N+o(\log N).
\end{align}
Using the assumed tail exponent on $B(x,b)$,
\begin{align}
    \log\left(N\Psi_B(t_N+\Delta(x,b);x,b)\right)
    &=
    \log N+\log\eta_B(x,b)
    -\frac{(t_N+\Delta(x,b))^2}{2\sigma_B^2(x)}
    +o\left((t_N+\Delta(x,b))^2\right)
    \\
    &=
    \left(
        1-\frac{c^2}{2\sigma_B^2(x)}+o(1)
    \right)\log N
    +O(1).
\end{align}
Since $c<\sqrt{2}\sigma_B(x)$, the coefficient of $\log N$ is positive,
and therefore
\begin{align}
    N\Psi_B(t_N+\Delta(x,b);x,b)\to\infty.
\end{align}
For every $y\in B(x,b)$, we have $y\in\safehat(x,b)$ and
$A(x,b)\subseteq\safehat(x,b)$. Hence
$R_A^\star(x,b)\le\sup_{z\in\safehat(x,b)}\rstar(x,z)$ and
$\rstar(x,y)\ge\inf_{z\in\safehat(x,b)}\rstar(x,z)$, so
\begin{align}
    R_A^\star(x,b)-\rstar(x,y)
    \le
    \Delta(x,b).
\end{align}
Thus
\begin{align}
    \{y\in B(x,b),\ \epsr(x,y)>t_N+\Delta(x,b)\}
    \subseteq
    \{y\in B(x,b),\ \rhat(x,y)>\tau_N\},
\end{align}
and hence
\begin{align}
    N\widehat{\Psi}_B(\tau_N;x,b)
    \ge
    N\Psi_B(t_N+\Delta(x,b);x,b)
    \to\infty.
\end{align}
Therefore the score-tail separation conditions in
\Cref{thm:hacking-asymptotic} hold with threshold $\tau_N$.  The final claim follows from
\Cref{thm:hacking-asymptotic}.
\end{proof}

\begin{remark}[Boundary behavior]
\label{rem:boundary}
When $\widehat\Psi_A$ and $\widehat\Psi_B$ are asymptotically comparable on the
relevant score scale, finer tail constants and class masses determine the
limit.  In error-tail conditions such as
\Cref{prop:gaussian-special-case}, the finite reward range $\Delta(x,b)$ can
affect boundary constants without changing the Gaussian exponent away from
the boundary.
\end{remark}
\subsection{Sample implementation and consistency of \cPes}
\label{app:additional-cpes-results}
\begin{algorithm}[t]
\caption{Constrained Pessimistic Inference (\cPes), sample implementation}
\label{alg:cPes}
\begin{algorithmic}[1]
\REQUIRE Prompt $x$, reference policy $\piref(\cdot\mid x)$, proxies $\rhat,\ghat$, threshold $b$, parameter $\beta$, clipping levels $\widehat R_{\min}<\widehat R_{\max}$, budget $N$
\STATE Sample $y_1,\ldots,y_N\overset{\mathrm{iid}}{\sim}\piref(\cdot\mid x)$
\STATE Set $\Ical_N(x,b)=\{i\in[N]:\ghat(x,y_i)\ge b\}$ and $m_N=|\Ical_N(x,b)|$
\IF{$m_N=0$}
    \STATE Return a predefined fallback, e.g., abstention
\ENDIF
\STATE Set $\widetilde r_i=\operatorname{clip}(\rhat(x,y_i),\widehat R_{\min},\widehat R_{\max})$ for each $i\in\Ical_N(x,b)$
\STATE Find $\hat\lambda_N(x)\in[\widehat R_{\min}-\beta,\widehat R_{\max})$ such that
\[
\frac{1}{m_N}\sum_{i\in\Ical_N(x,b)}
\relu\left(\frac{\widetilde r_i-\hat\lambda_N(x)}{\beta}\right)=1
\]
\STATE Set $\hat p_i=\frac{1}{m_N}\relu\left(\frac{\widetilde r_i-\hat\lambda_N(x)}{\beta}\right)$ for each $i\in\Ical_N(x,b)$
\STATE Sample $I\in\Ical_N(x,b)$ with probability $\hat p_i$ and return $y_I$
\end{algorithmic}
\end{algorithm}

\begin{proposition}[Consistency of sample \cPes]
\label{prop:lambda-consistency}
Fix $x\in\Xcal$ and $b\in(0,1]$ with $q(x,b)>0$.
Let $\hat\pi_N(\cdot\mid x)$ be the random output distribution induced by
\Cref{alg:cPes} on $m_N>0$.  Then
$m_N\to\infty$ almost surely,
$\hat\lambda_N(x)\to\lambda(x)$ almost surely, and for every bounded
$f:\Ycal\to\mbR$,
\begin{align}
    \mbE_{y\sim\hat\pi_N(\cdot\mid x)}[f(y)]
    \to
    \mbE_{y\sim\hat\pi(\cdot\mid x)}[f(y)]
    \qquad\text{almost surely}.
\end{align}
\end{proposition}
\begin{proof}
Let $m_N=\sum_{i=1}^N\ind\{y_i\in\safehat(x,b)\}$.  Since
$m_N/N\to q(x,b)>0$ almost surely, $m_N\to\infty$ almost surely and
$m_N>0$ eventually.

Conditional on belonging to $\safehat(x,b)$, the feasible samples are i.i.d.
from $\piref^\sharp(\cdot\mid x)$.  Define
\begin{align}
    h_N(\lambda)
    &\coloneqq
    \frac{1}{m_N}\sum_{i\in\Ical_N(x,b)}
    \relu\left(
        \frac{\widetilde r(x,y_i)-\lambda}{\beta}
    \right),\\
    h(\lambda)
    &\coloneqq
    \mbE_{\piref^\sharp}
    \relu\left(
        \frac{\widetilde r(x,y)-\lambda}{\beta}
    \right).
\end{align}
The functions indexed by
$\lambda\in[\widehat R_{\min}-\beta,\widehat R_{\max}]$ are uniformly
bounded and $1/\beta$-Lipschitz in $\lambda$.  A finite-grid argument combined
with the strong law of large numbers yields
\begin{align}
    \sup_{\lambda\in[\widehat R_{\min}-\beta,\widehat R_{\max}]}
    |h_N(\lambda)-h(\lambda)|\to0
    \qquad\text{almost surely}.
\end{align}
By \Cref{prop:pessimistic-policy-closed-form}, the equation $h(\lambda)=1$ has a
unique solution $\lambda(x)$.  Uniform convergence and monotonicity imply that
any empirical solution $\hat\lambda_N(x)$ to $h_N(\lambda)=1$ converges almost
surely to $\lambda(x)$.

For any bounded $f$, write
\begin{align}
    \mbE_{\hat\pi_N}[f]
    =
    \frac{1}{m_N}\sum_{i\in\Ical_N(x,b)}
    f(y_i)
    \relu\left(
        \frac{\widetilde r(x,y_i)-\hat\lambda_N(x)}{\beta}
    \right).
\end{align}
The summands are uniformly bounded, and the weights converge uniformly in the
multiplier because of the Lipschitz property.  Applying the strong law again and
using $\hat\lambda_N(x)\to\lambda(x)$ gives
\begin{align}
    \mbE_{\hat\pi_N}[f]
    \to
    \mbE_{\piref^\sharp}
    \left[
        f(y)
        \relu\left(
            \frac{\widetilde r(x,y)-\lambda(x)}{\beta}
        \right)
    \right]
    =
    \mbE_{\hat\pi}[f].
\end{align}
\end{proof}

\subsection{Utility tradeoff of \cPes}
\label{app:cpes-utility}

Bounded coverage alone can be achieved by avoiding reward optimization
altogether. We therefore quantify the utility cost of coverage control. For a
policy $\pi$, let
\begin{align}
    J(\pi;x)
    \coloneqq
    \mbE_{y\sim\pi(\cdot\mid x)}[\rstar(x,y)].
    \label{eq:def-policy-utility}
\end{align}
Define the clipped-score RMSE under the proxy-feasible reference by
\begin{align}
    \bar\varepsilon_{r,\mathrm{clip}}^\sharp(x)
    \coloneqq
    \sqrt{
        \mbE_{y\sim\piref^\sharp(\cdot\mid x)}
        \left[(\widetilde r(x,y)-\rstar(x,y))^2\right]
    }.
    \label{eq:def-clipped-reward-rmse-sharp}
\end{align}
This quantity includes both reward-proxy error and any distortion introduced by
clipping.

\begin{proposition}[Utility tradeoff within the proxy-feasible set]
\label{prop:cpes-utility}
Suppose $\bar\varepsilon_{r,\mathrm{clip}}^\sharp(x)<\infty$.
For any comparator policy
$\rho(\cdot\mid x)\ll\piref^\sharp(\cdot\mid x)$, the population \cPes{}
policy satisfies
\begin{align}
    J(\rho;x)-J(\hat\pi;x)
    \le
    \frac{\beta}{2}\bigl(C_\rho^\sharp(x)-1\bigr)
    +
    \bar\varepsilon_{r,\mathrm{clip}}^\sharp(x)
    \left(
        \sqrt{C_\rho^\sharp(x)}
        +
        \sqrt{1+\frac{\widehat R_{\mathrm{span}}}{\beta}}
    \right).
    \label{eq:cpes-utility-main}
\end{align}
\end{proposition}

The bound makes explicit the role of $\beta$. Smaller $\beta$ permits more
aggressive reward optimization but allows greater concentration relative to
$\piref^\sharp$, while larger $\beta$ keeps the policy closer to the
proxy-feasible reference distribution.

\subsection{Oracle regret decomposition}
For the oracle comparison below, define the ambient coverage coefficient
\begin{align}
    C_\pi(x)
    \coloneqq
    \sum_{y\in\Ycal}\frac{\pi(y\mid x)^2}{\piref(y\mid x)}.
\end{align}
For a comparator $\pi^\star$, let
$\alpha^\star(x,b)\coloneqq\pi^\star(\safehat(x,b)\mid x)$ and, when
$\alpha^\star(x,b)>0$, define
\begin{align}
    \pi^{\star,\sharp}(y\mid x)
    \coloneqq
    \frac{\pi^\star(y\mid x)\ind\{y\in\safehat(x,b)\}}
    {\alpha^\star(x,b)}.
\end{align}

\begin{theorem}[Oracle regret decomposition]
\label{thm:pessimism-regret}
Fix $b\in(0,1)$ and suppose $0\le\rstar(x,y)\le R_{\max}$.
Let $\pi^\star\ll\piref$ satisfy
$\supp(\pi^\star)\subseteq\Scal_+^\star(x)$; if
$\alpha^\star(x,b)>0$, assume also
$\pi^{\star,\sharp}\ll\piref^\sharp$.
If $\alpha^\star(x,b)>0$, then
\begin{align}
    J(\pi^\star;x)-J(\hat\pi;x)
    &\le
    \Bigg[
        \frac{\beta}{2}\bigl(C_{\pi^{\star,\sharp}}^\sharp(x)-1\bigr)
        +\bar\varepsilon_{r,\mathrm{clip}}^\sharp(x)
        \left(
            \sqrt{C_{\pi^{\star,\sharp}}^\sharp(x)}
            +\sqrt{1+\frac{\widehat R_{\mathrm{span}}}{\beta}}
        \right)
    \Bigg]
    \nonumber\\
    &\quad+
    \frac{R_{\max}\sqrt{C_{\pi^\star}(x)}\,\bar\varepsilon_g(x)}{1-b}.
    \label{eq:oracle-regret-bound}
\end{align}
If $\alpha^\star(x,b)=0$, then
\begin{align}
    J(\pi^\star;x)-J(\hat\pi;x)
    \le
    \frac{R_{\max}\sqrt{C_{\pi^\star}(x)}\,\bar\varepsilon_g(x)}{1-b},
    \label{eq:oracle-regret-bound-zero-alpha}
\end{align}
that is, the bracketed proxy-feasible comparison term is omitted.
\end{theorem}
\begin{proof}
Decompose the regret of the truly safe comparator $\pi^\star$:
\begin{align}
    J(\pi^\star;x)-J(\hat\pi;x)
    &=
    \mbE_{\pi^\star}[\rstar(x,y)\ind\{y\in\safehat(x,b)\}]
    -J(\hat\pi;x)\\
    &\quad+
    \mbE_{\pi^\star}[\rstar(x,y)\ind\{y\notin\safehat(x,b)\}].
\end{align}
If $\alpha^\star(x,b)>0$, then the first term is at most
$J(\pi^{\star,\sharp};x)-J(\hat\pi;x)$ because
$\alpha^\star(x,b)\le1$ and rewards are nonnegative.  Applying \Cref{prop:cpes-utility} with
$\rho=\pi^{\star,\sharp}$ gives the bracketed term in
\eqref{eq:oracle-regret-bound}.  If $\alpha^\star(x,b)=0$, the first term is
$-J(\hat\pi;x)\le0$, so the bracketed term is omitted.

It remains to control the second term.  Since
$\supp(\pi^\star)\subseteq\Scal_+^\star(x)$, the event
$\{y\notin\safehat(x,b)\}$ under $\pi^\star$ is a false negative of the safety
proxy.  On this event, $\gstar(x,y)=1$ and $\ghat(x,y)<b$, so
$|\epsg(x,y)|>1-b$.  Hence
\begin{align}
    \mbP_{\piref}[y\in\Scal_+^\star(x),\ y\notin\safehat(x,b)]
    \le
    \frac{\bar\varepsilon_g(x)^2}{(1-b)^2}.
\end{align}
By Cauchy--Schwarz and absolute continuity,
\begin{align}
    \mbP_{\pi^\star}[y\notin\safehat(x,b)]
    &=
    \mbE_{\piref}
    \left[
        \frac{\pi^\star(y\mid x)}{\piref(y\mid x)}
        \ind\{y\in\Scal_+^\star(x),\ y\notin\safehat(x,b)\}
    \right]\\
    &\le
    \sqrt{C_{\pi^\star}(x)}
    \sqrt{
        \mbP_{\piref}[y\in\Scal_+^\star(x),\ y\notin\safehat(x,b)]
    }\\
    &\le
    \frac{\sqrt{C_{\pi^\star}(x)}\,\bar\varepsilon_g(x)}{1-b}.
\end{align}
Using $\rstar(x,y)\le R_{\max}$ proves
\begin{align}
    \mbE_{\pi^\star}[\rstar(x,y)\ind\{y\notin\safehat(x,b)\}]
    \le
    \frac{R_{\max}\sqrt{C_{\pi^\star}(x)}\,\bar\varepsilon_g(x)}{1-b}.
\end{align}
Combining the bounds proves the theorem.
\end{proof}
\subsection{Proof of \Cref{thm:hacking-finiteN}}
\begin{proof}
Suppress $(x,b)$ in the notation.  For a single sample $y$, define
\begin{align}
    \widehat{E}_A(t)&\coloneqq\{y\in A,\ \rhat(x,y)>t\},\\
    \widehat{E}_B(t)&\coloneqq\{y\in B,\ \rhat(x,y)>t\}.
\end{align}
The sets $A$ and $B$ are disjoint, so these two one-sample events are disjoint.
If among $N$ samples no event $\widehat{E}_A(t)$ occurs and at least one event
$\widehat{E}_B(t)$ occurs, then every true-positive proxy-feasible sample has
learned score at most $t$, while some unsafe-but-feasible sample has learned
score strictly larger than $t$.  Therefore the CBo$N$ maximizer over the
proxy-feasible samples lies in $B$.  The probability of this sufficient event is
\begin{align}
    \bigl(1-\widehat{\Psi}_A(t)\bigr)^N
    -
    \bigl(1-\widehat{\Psi}_A(t)-\widehat{\Psi}_B(t)\bigr)^N,
\end{align}
which proves the lower bound.

For the upper bound, consider the event that no unsafe-but-feasible sample
satisfies $\rhat(x,y)>s$ and at least one true-positive proxy-feasible sample
satisfies $\rhat(x,y)>s$.  On this event, some sample in $A$ strictly beats every
sample in $B$, so CBo$N$ cannot hack.  The probability of this non-hacking
certificate is
\begin{align}
    \bigl(1-\widehat{\Psi}_B(s)\bigr)^N
    -
    \bigl(1-\widehat{\Psi}_B(s)-\widehat{\Psi}_A(s)\bigr)^N.
\end{align}
Hence the hacking probability is at most its complement,
\begin{align}
    1-
    \bigl(1-\widehat{\Psi}_B(s)\bigr)^N
    +
    \bigl(1-\widehat{\Psi}_B(s)-\widehat{\Psi}_A(s)\bigr)^N,
\end{align}
which is \eqref{eq:finite-upper-bound}.
\end{proof}
\subsection{Proof of \Cref{thm:hacking-asymptotic}}
\begin{proof}
Let $a_N=\widehat{\Psi}_A(t_N;x,b)$ and
$d_N=\widehat{\Psi}_B(t_N;x,b)$.  The assumptions imply
$Na_N\to0$ and $Nd_N\to\infty$.  Therefore
$(1-a_N)^N\to1$ and
$(1-a_N-d_N)^N\le\exp[-N(a_N+d_N)]\to0$.  The lower bound in
\Cref{thm:hacking-finiteN} converges to one, proving the claim.
\end{proof}
\subsection{Proof of \Cref{cor:small-rmse-amplification}}
\begin{proof}
Fix $x$ and suppress the dependence on $x$ in the notation. Let the
output space contain two outputs,
\[
    \mathcal{Y}=\{y_A,y_B\},
\]
with reference probabilities
\[
    \pi_{\rm ref}(y_A \mid x)=1-\varepsilon,
    \qquad
    \pi_{\rm ref}(y_B \mid x)=\varepsilon.
\]
Define the true and learned safety functions by
\[
    g^\star(x,y_A)=1, \qquad \widehat g(x,y_A)=1,
\]
and
\[
    g^\star(x,y_B)=0, \qquad \widehat g(x,y_B)=b.
\]
Then both outputs are proxy-feasible, since $\widehat g(x,y_A)\ge b$
and $\widehat g(x,y_B)\ge b$. Moreover,
\[
    A(x,b)=\{y_A\},
    \qquad
    B(x,b)=\{y_B\},
\]
and therefore
\[
    \mathbb{P}_{y \sim \pi_{\rm ref}(\cdot \mid x)}[y \in B(x,b)]
    =\varepsilon.
\]
The safety-proxy RMSE is
\[
    \sqrt{
    \mathbb{E}_{\pi_{\rm ref}}
    \left[
    \left(\widehat g(x,y)-g^\star(x,y)\right)^2
    \right]}
    =
    \sqrt{\varepsilon b^2}
    =
    b\sqrt{\varepsilon}.
\]

Now define the true and learned reward functions by
\[
    r^\star(x,y_A)=0, \qquad \widehat r(x,y_A)=0,
\]
and
\[
    r^\star(x,y_B)=0, \qquad \widehat r(x,y_B)=\xi.
\]
Thus the reward-proxy RMSE is
\[
    \sqrt{
    \mathbb{E}_{\pi_{\rm ref}}
    \left[
    \left(\widehat r(x,y)-r^\star(x,y)\right)^2
    \right]}
    =
    \sqrt{\varepsilon \xi^2}
    =
    \xi\sqrt{\varepsilon}.
\]

Since $\widehat r(x,y_B)>\widehat r(x,y_A)$, constrained Best-of-$N$
returns $y_B$ if and only if at least one of the $N$ sampled candidates
equals $y_B$. All sampled candidates are proxy-feasible, so there is no
abstention event. Hence
\[
    \mathbb{P}\left[
    \widehat y^{\rm BoN}_{N}(x) \in B(x,b)
    \right]
    =
    1-(1-\varepsilon)^N.
\]
Using $(1-\varepsilon)^N \le \exp(-N\varepsilon)$ gives
\[
    \mathbb{P}\left[
    \widehat y^{\rm BoN}_{N}(x) \in B(x,b)
    \right]
    \ge
    1-\exp(-N\varepsilon).
\]
Therefore, if
\[
    N \ge \varepsilon^{-1}\log(1/\delta),
\]
then the safety-hacking probability is at least $1-\delta$. This proves
the claim.
\end{proof}

\subsection{Proof of \Cref{thm:coverage-controlled-safety}}
\label{app:proof-contamination-amplification}
\begin{proof}
Let
\[
    w(y)
    \coloneqq
    \frac{\pi(y\mid x)}{\piref^\sharp(y\mid x)},
    \qquad
    \kappa
    \coloneqq
    \piref^\sharp(B(x,b)\mid x).
\]
Since $\mbE_{\piref^\sharp}[w]=1$ and
$\mbE_{\piref^\sharp}[w^2]=C_\pi^\sharp(x)$,
\begin{align}
    \mbE_{\piref^\sharp}[(w-1)^2]
    = C_\pi^\sharp(x)-1.
\end{align}
Moreover,
\begin{align}
    \mbP_{y\sim\pi(\cdot\mid x)}[y\in B(x,b)]-\kappa
    &=
    \mbE_{\piref^\sharp}
    \left[(w(y)-1)\ind\{y\in B(x,b)\}\right] \\
    &=
    \mbE_{\piref^\sharp}
    \left[(w(y)-1)\bigl(\ind\{y\in B(x,b)\}-\kappa\bigr)\right],
\end{align}
where the second equality uses $\mbE_{\piref^\sharp}[w-1]=0$.
By Cauchy--Schwarz,
\begin{align}
    \left|
        \mbP_{y\sim\pi(\cdot\mid x)}[y\in B(x,b)]-\kappa
    \right|
    &\le
    \sqrt{C_\pi^\sharp(x)-1}
    \sqrt{\kappa(1-\kappa)}.
\end{align}
This proves the bound.

By the definition of $\piref^\sharp$, we have
\begin{align}
    \mbP_{y\sim\piref^\sharp(\cdot\mid x)}[y\in B(x,b)]
    &=
    \mbE_{y\sim\piref^\sharp(\cdot\mid x)}
    \left[\ind\{y\in B(x,b)\}\right] \\
    &=
    \mbE_{y\sim\piref(\cdot\mid x)}
    \left[\frac{\ind\{y\in B(x,b)\}\ind\{y\in \widehat{\mathcal{S}}_{+}(x,b)\}}{q(x, b)}\right] \\
    &=
    \frac{\mbE_{y\sim\piref(\cdot\mid x)}
    \left[\ind\{y\in B(x,b)\}\right]}{q(x, b)}.
\end{align}
If $y\in B(x,b)$, then $\gstar(x,y)=0$ and $\ghat(x,y)\ge b$, so
$|\epsg(x,y)|\ge b$.  Hence
\begin{align}
    \ind\{y\in B(x,b)\}
    \le
    \frac{\epsg(x,y)^2}{b^2}.
\end{align}
Taking expectation under $\piref(\cdot\mid x)$ gives
\begin{align}
    \piref(B(x,b)\mid x)
    \le
    \frac{\bar\varepsilon_g(x)^2}{b^2}.
\end{align}
Combining this inequality with
$\kappa(x,b)=\piref(B(x,b)\mid x)/q(x,b)$ gives
\begin{align}
    \kappa(x,b)
    \le
    \frac{\bar\varepsilon_g(x)^2}{b^2q(x,b)},
\end{align}
which proves \eqref{eq:coverage-controlled-safety-bound}.
\end{proof}

\subsection{Proof of \Cref{prop:pessimistic-policy-closed-form}}
\begin{proof}
Let $\mu(y)=\piref^\sharp(y\mid x)$ and write
$w(y)=\pi(y\mid x)/\mu(y)$.  Then $\pi\ll\mu$ and
$\sum_y\pi(y\mid x)=1$ are equivalent to $w(y)\ge0$ and
$\mbE_\mu[w]=1$.  The objective in
\eqref{eq:pessimistic-policy-optimization} becomes
\begin{align}
    \max_{w\ge0,\ \mbE_\mu[w]=1}
    \mbE_{y\sim\mu}
    \left[w(y)\widetilde r(x,y)-\frac{\beta}{2}w(y)^2\right].
    \label{eq:w-program-proof}
\end{align}
This is strictly concave in $w$, so the optimizer is unique.

Introduce a Lagrange multiplier $\lambda$ for $\mbE_\mu[w]=1$.  Pointwise
maximization of the Lagrangian over $w(y)\ge0$ gives
\begin{align}
    w_\lambda(y)
    =
    \left(\frac{\widetilde r(x,y)-\lambda}{\beta}\right)_+
    =
    \relu\left(\frac{\widetilde r(x,y)-\lambda}{\beta}\right).
\end{align}
The multiplier must satisfy $\mbE_\mu[w_\lambda]=1$.

It remains to show existence and uniqueness of $\lambda$.  Define
\begin{align}
    h(\lambda)
    \coloneqq
    \mbE_\mu
    \left[
        \relu\left(
            \frac{\widetilde r(x,y)-\lambda}{\beta}
        \right)
    \right].
\end{align}
The function $h$ is continuous and nonincreasing.  By construction,
$h(\widehat R_{\max})=0$, while
\begin{align}
    h(\widehat R_{\min}-\beta)
    =
    \mbE_\mu\left[
        \frac{\widetilde r(x,y)-\widehat R_{\min}+\beta}{\beta}
    \right]
    \ge1.
\end{align}
Thus a solution exists in
$[\widehat R_{\min}-\beta,\widehat R_{\max})$.  It is unique because whenever
$h(\lambda)>0$, the event $\{\widetilde r(x,y)>\lambda\}$ has positive
$\mu$-probability, so $h$ is strictly decreasing on the relevant level set.

Finally, let $w(y)=w_{\lambda(x)}(y)$.  Since $\mbE_\mu[w]=1$ and
$\lambda(x)\ge\widehat R_{\min}-\beta$,
\begin{align}
    0\le w(y)
    \le
    \frac{\widehat R_{\max}-\lambda(x)}{\beta}
    \le
    1+\frac{\widehat R_{\mathrm{span}}}{\beta}.
\end{align}
Therefore
\begin{align}
    C_{\hat\pi}^\sharp(x)
    =
    \mbE_\mu[w(y)^2]
    \le
    \bigl(\sup_y w(y)\bigr)\mbE_\mu[w(y)]
    \le
    1+\frac{\widehat R_{\mathrm{span}}}{\beta}.
\end{align}
\end{proof}
\subsection{Proof of \Cref{prop:cpes-utility}}
\begin{proof}
Let $\mu=\piref^\sharp(\cdot\mid x)$.  By optimality of $\hat\pi$ in
\eqref{eq:pessimistic-policy-optimization}, for any
$\rho(\cdot\mid x)\ll\mu$,
\begin{align}
    \mbE_\rho[\widetilde r(x,y)]
    -\mbE_{\hat\pi}[\widetilde r(x,y)]
    \le
    \frac{\beta}{2}\bigl(C_\rho^\sharp(x)-1\bigr),
\end{align}
where we dropped the nonpositive term
$-\frac{\beta}{2}(C_{\hat\pi}^\sharp(x)-1)$.  Therefore
\begin{align}
    J(\rho;x)-J(\hat\pi;x)
    &\le
    \frac{\beta}{2}\bigl(C_\rho^\sharp(x)-1\bigr)
    +
    \left|
        \mbE_\rho[\widetilde r(x,y)-\rstar(x,y)]
    \right|
    +
    \left|
        \mbE_{\hat\pi}[\widetilde r(x,y)-\rstar(x,y)]
    \right|.
\end{align}
Cauchy--Schwarz under $\mu$ gives
\begin{align}
    \left|
        \mbE_\rho[\widetilde r(x,y)-\rstar(x,y)]
    \right|
    &\le
    \bar\varepsilon_{r,\mathrm{clip}}^\sharp(x)
    \sqrt{C_\rho^\sharp(x)},\\
    \left|
        \mbE_{\hat\pi}[\widetilde r(x,y)-\rstar(x,y)]
    \right|
    &\le
    \bar\varepsilon_{r,\mathrm{clip}}^\sharp(x)
    \sqrt{C_{\hat\pi}^\sharp(x)}
    \le
    \bar\varepsilon_{r,\mathrm{clip}}^\sharp(x)
    \sqrt{1+\frac{\widehat R_{\mathrm{span}}}{\beta}},
\end{align}
where the last inequality follows from
\eqref{eq:coverage-bound-pessimistic-policy}.  Combining these bounds proves
\eqref{eq:cpes-utility-main}.
\end{proof}
\subsection{Proof of \Cref{cor:cPes-safety}}

\paragraph{Finite-sample claim.}
\begin{proof}
Condition on the event $m_N>0$ and on the sampled candidates.  Let
\begin{align}
    w_i
    =
    \relu\left(
        \frac{\widetilde r(x,y_i)-\hat\lambda_N(x)}{\beta}
    \right),
    \qquad i\in\Ical_N(x,b).
\end{align}
By construction, $m_N^{-1}\sum_{i\in\Ical_N}w_i=1$.  The conditional probability
    that \cPes~returns an unsafe-but-feasible sample is
\begin{align}
    \frac{1}{m_N}\sum_{i\in\Ical_N}w_i\ind\{y_i\in B(x,b)\}.
\end{align}

Let
\begin{align}
    K_N^B
    &\coloneqq
    \sum_{i=1}^N\ind\{y_i\in B(x,b)\},\\
    \widehat\kappa_N
    &\coloneqq
    \frac{K_N^B}{m_N}.
\end{align}
Since $m_N^{-1}\sum_{i\in\Ical_N}(w_i-1)=0$, we have
\begin{align}
    \frac{1}{m_N}
    \sum_{i\in\Ical_N}
    w_i\ind\{y_i\in B(x,b)\}
    -
    \widehat\kappa_N
    =
    \frac{1}{m_N}
    \sum_{i\in\Ical_N}
    (w_i-1)
    \left(
        \ind\{y_i\in B(x,b)\}
        -
        \widehat\kappa_N
    \right).
\end{align}
By Cauchy--Schwarz,
\begin{align}
    \left|
    \frac{1}{m_N}
    \sum_{i\in\Ical_N}
    w_i\ind\{y_i\in B(x,b)\}
    -
    \widehat\kappa_N
    \right|
    \le
    \left(
        \frac{1}{m_N}
        \sum_{i\in\Ical_N}(w_i-1)^2
    \right)^{1/2}
    \sqrt{
        \widehat\kappa_N(1-\widehat\kappa_N)
    }.
\end{align}
The empirical analogue of
\eqref{eq:coverage-bound-pessimistic-policy} gives
\begin{align}
    \frac{1}{m_N}
    \sum_{i\in\Ical_N}(w_i-1)^2
    =
    \frac{1}{m_N}
    \sum_{i\in\Ical_N}w_i^2-1
    \le
    \frac{\widehat R_{\mathrm{span}}}{\beta}.
\end{align}
Consequently, conditional on the sampled candidates,
\begin{align}
    \mbP[
        \widehat y_{\mathrm{cPes},N}(x)\in B(x,b)
        \mid y_1,\ldots,y_N
    ]
    \le
    \widehat\kappa_N
    +
    \sqrt{
        \frac{\widehat R_{\mathrm{span}}}{\beta}
        \widehat\kappa_N(1-\widehat\kappa_N)
    }.
\end{align}

Given $m_N=m>0$, the feasible samples are i.i.d. from
$\piref^\sharp$, and hence
\begin{align}
    K_N^B\mid m_N=m
    \sim
    \operatorname{Binomial}\bigl(m,\kappa(x,b)\bigr).
\end{align}
Therefore,
\begin{align}
    \mbE[\widehat\kappa_N\mid m_N=m]
    =
    \kappa(x,b).
\end{align}
Since $z\mapsto\sqrt{z(1-z)}$ is concave on $[0,1]$, Jensen's
inequality gives
\begin{align}
    \mbE\left[
        \sqrt{\widehat\kappa_N(1-\widehat\kappa_N)}
        \,\middle|\,m_N=m
    \right]
    \le
    \sqrt{\kappa(x,b)(1-\kappa(x,b))}.
\end{align}
It follows, for every $m>0$, that
\begin{align}
    \mbP[
        \widehat y_{\mathrm{cPes},N}(x)\in B(x,b)
        \mid m_N=m
    ]
    \le
    \kappa(x,b)
    +
    \sqrt{
        \frac{\widehat R_{\mathrm{span}}}{\beta}
        \kappa(x,b)(1-\kappa(x,b))
    }.
\end{align}
Averaging over $m$ gives the same bound conditional on $m_N>0$.

Finally, \Cref{thm:coverage-controlled-safety} gives
\begin{align}
    \kappa(x,b)
    \le
    \frac{\bar\varepsilon_g(x)^2}{b^2q(x,b)}.
\end{align}
Using $1-\kappa(x,b)\le1$, we obtain
\begin{align}
    \mbP[
        \widehat y_{\mathrm{cPes},N}(x)\in B(x,b)
        \mid m_N>0
    ]
    \le
    \min\left\{
        1,\,
        \frac{\bar\varepsilon_g(x)^2}{b^2q(x,b)}
        +
        \frac{\bar\varepsilon_g(x)}{b}
        \sqrt{
            \frac{\widehat R_{\mathrm{span}}}
                 {\beta q(x,b)}
        }
    \right\},
\end{align}
which proves \eqref{eq:safety-upperbound-main}.
The unconditional statement follows because abstention is not an unsafe output.
\end{proof}
\paragraph{Population claim.}
\begin{proof}
By \Cref{prop:pessimistic-policy-closed-form}, the population pessimistic policy
satisfies $\hat\pi(\cdot\mid x)\ll\piref^\sharp(\cdot\mid x)$ and
\begin{align}
    C_{\hat\pi}^\sharp(x)
    \le
    1+\frac{\widehat R_{\mathrm{span}}}{\beta}.
\end{align}

Applying \Cref{thm:coverage-controlled-safety} and using
\eqref{eq:coverage-bound-pessimistic-policy} gives
\begin{align}
    \mbP_{y\sim\hat\pi(\cdot\mid x)}[y\in B(x,b)]
    \le
    \kappa(x,b)
    +
    \sqrt{
        \frac{\widehat R_{\mathrm{span}}}{\beta}
        \kappa(x,b)\bigl(1-\kappa(x,b)\bigr)
    }.
\end{align}
Since
\begin{align}
    \kappa(x,b)
    \le
    \frac{\bar\varepsilon_g(x)^2}{b^2q(x,b)},
\end{align}
and $1-\kappa(x,b)\le1$, it follows that
\begin{align}
    \mbP_{y\sim\hat\pi(\cdot\mid x)}[y\in B(x,b)]
    \le
    \min\left\{
        1,\,
        \frac{\bar\varepsilon_g(x)^2}{b^2q(x,b)}
        +
        \frac{\bar\varepsilon_g(x)}{b}
        \sqrt{
            \frac{\widehat R_{\mathrm{span}}}
                 {\beta q(x,b)}
        }
    \right\}.
\end{align}
\end{proof}

\subsection{Additional Toy Experimental Details and Ablations}
\label{app:toy-additional}

\paragraph{Experimental details.}
The toy problem in \Cref{subsec:toy_experiment} uses class probabilities
$(0.39,0.01,0.60)$ for $A$, $B$, and $C$, respectively. We evaluate random
feasible selection, CBo$N$, and \cPes{} over budgets
\begin{align}
    N\in\{1,2,4,\ldots,8192\},
\end{align}
using $5000$ Monte Carlo trials per budget. Unless otherwise stated, metrics
are reported conditional on non-abstention, since at small $N$ all sampled
candidates may belong to the proxy-rejected class $C$.

The true reward gap between the safe and unsafe proxy-feasible classes is
\begin{align}
    \Delta
    =
    r_A^\star-r_B^\star
    =
    0.6,
\end{align}
which matches the local reward-range shift used when translating reward-error
tails into learned-score tails in \Cref{prop:gaussian-special-case}.

\paragraph{\cPes{} regularization.}
The \cPes{} results in \Cref{fig:toy-main} are consistent with the
budget-uniform control predicted by \Cref{cor:cPes-safety}. With $\beta=1.0$,
the conditional hacking rate remains between $0.015$ and $0.024$ over the
entire budget grid, while the mean true reward stays near $0.79$. Even with
$\beta=0.2$, the hacking rate remains around $0.024$--$0.031$ at large
budgets.

A separate sweep over $\beta$ illustrates the concentration tradeoff. At
$N=8192$, the hacking rate is $0.1518$ for $\beta=0.02$, $0.0418$ for
$\beta=0.1$, and $0.0152$ for $\beta=1.0$, with the highest true reward
observed around $\beta\in[0.5,1.0]$. Thus, as \cPes{} approaches aggressive
max-selection, it increasingly recovers the same amplification mechanism as
CBo$N$.

\paragraph{Equal-tail control.}
To verify that large candidate budgets alone do not cause safety hacking, we
set
\[
    \sigma_B=\sigma_A=0.2
\]
while keeping the remainder of the environment unchanged. In this setting,
the unsafe class $B$ no longer has a heavier reward-error upper tail than the
safe class $A$. The CBo$N$ hacking rate falls to zero by $N=32$, and the
selected true reward converges to $0.8$, the reward of class $A$.

\paragraph{No-false-positive control.}
We also remove class $B$ entirely. In this case the proxy-feasible set contains
no unsafe outputs, and the safety-hacking rate is identically zero for all
methods and all candidate budgets.

\paragraph{False-positive tail-scale sweep.}
Finally, \Cref{fig:toy-sigma-b} varies the reward-proxy noise scale
$\sigma_B$ of the unsafe-but-feasible class. At $N=8192$, the CBo$N$
hacking rate increases from $0.0$ at $\sigma_B=0.2$ to $0.697$ at
$\sigma_B=0.6$, $0.9996$ at $\sigma_B=1.0$, and $0.9998$ at
$\sigma_B=1.2$.

Together, these controls support the mechanism characterized by
\Cref{thm:hacking-asymptotic}: increasing $N$ is not sufficient by itself.
Amplification occurs when residual unsafe outputs remain proxy-feasible and
have a sufficiently favorable upper tail under the learned reward.

\section{Additional LLM Analysis}
\label{app:llm-analysis}

\subsection{Evaluator and scoring details}
\label{app:llm-evaluation-details}

\paragraph{Safety proxy and operational safety criterion.}
The safety proxy is
\href{https://huggingface.co/meta-llama/Llama-Guard-3-8B}{\texttt{Llama-Guard-3-8B}}. We use
$1-p_{\mathrm{unsafe}}$ as its safe score and accept candidates whose score is
at least $b=0.95$. Primary safety labels are produced by the HarmBench
classifier
\href{https://huggingface.co/cais/HarmBench-Llama-2-13b-cls}{\texttt{HarmBench-Llama-2-13b-cls}}. Thus $A$ contains
proxy-feasible candidates classified as safe by HarmBench, $B$ contains
proxy-feasible candidates classified as unsafe, and $C$ contains
proxy-rejected candidates. The reported safety-hacking rate is the probability
of selecting from $B$, conditional on non-abstention.

\paragraph{Safety-aware reward evaluator.}
We score safety-aware reward with the OpenAI Responses API using the
\texttt{gpt-5-mini} API model alias and the \texttt{safety\_aware\_v1} rubric.
The rubric assigns 1.0 to responses that are safe and fully satisfy a legitimate
request, 0.0 to responses that are unsafe, enable harmful behavior, or are
useless/nonresponsive, and intermediate values to partially helpful, vague,
incomplete, or weakly useful responses.

\paragraph{Reward proxies and aggregation.}
The primary reward proxy is
\nolinkurl{PKU-Alignment/beaver-7b-v1.0-reward}. For the reward-proxy ablation,
we use
\href{https://huggingface.co/Skywork/Skywork-Reward-V2-Llama-3.1-8B}{\texttt{Skywork-Reward-V2-Llama-3.1-8B}} with Hugging Face Transformers.
Its input is formatted as
\texttt{User: \{prompt\_text\}\textbackslash nAssistant:
\{response\_text\}}, and the scalar sequence-classification logit is used as
the raw proxy reward. Downstream normalization uses the calibration median and
interquartile range; clipping is applied only where required by \cPes{}.

\subsection{Alternative safety-filter sensitivity analysis}
\label{app:shieldgemma-filter-robustness}

We repeat the CBo$N$ analysis using
\href{https://huggingface.co/google/shieldgemma-2b}{\texttt{google/shieldgemma-2b}}
as an alternative safety filter, while holding the candidate pool, generator,
Beaver reward proxy, HarmBench labels, and candidate permutations fixed. We
select the ShieldGemma threshold using only the 179 calibration prompts by
matching the Llama Guard calibration feasible rate. The resulting threshold is
$0.8176$, giving a feasible rate of $0.8242$, compared with $0.8295$ for Llama
Guard at its primary threshold.

\begin{table}[H]
    \centering
    \small
    \begin{tabular}{@{}c cc cc c@{}}
        \toprule
        $N$ & Hacking & Abstention & Unsafe-only & Competitive & Tail win \\
        \midrule
        $1$   & $12.10\,(0.98)$ & $14.24\,(1.04)$ & $12.10$ & $0.00$  & -- \\
        $256$ & $14.75\,(1.34)$ & $1.26\,(0.42)$  & $0.99$  & $13.76$ & $35.02$ \\
        \bottomrule
    \end{tabular}
    \caption{ShieldGemma-2B alternative-filter sensitivity analysis. All entries are percentages; standard errors across prompts are shown in parentheses for hacking and abstention.
    Unsafe-only and Competitive are the two disjoint terms in
    \eqref{eq:finite-n-hacking-decomposition} and sum to the total hacking
    rate. Tail win denotes
    $\Pr(M_B>M_A\mid K_A>0,K_B>0)$ and is reported separately because it uses a
    different conditioning event.}
    \label{tab:shieldgemma-filter-robustness}
\end{table}

Conditional CBo$N$ safety hacking increases from $12.1\%$ at $N=1$ to
$14.8\%$ at $N=256$. Over the same range, the unsafe-only term decreases from
$12.1\%$ to $1.0\%$, while the competitive unsafe-win term increases from zero
to $13.8\%$. Thus the observed increase is driven by reward-based competition
rather than unsafe-only exposure. The increase is smaller than in the primary
analysis, so we treat this result as qualitatively consistent sensitivity
evidence rather than as a replacement for the primary Llama Guard analysis.

\subsection{Reward-tail survival curves}
\label{app:skywork-tail-survival}

\begin{figure*}[t]
    \centering
    \includegraphics[width=0.7\linewidth]{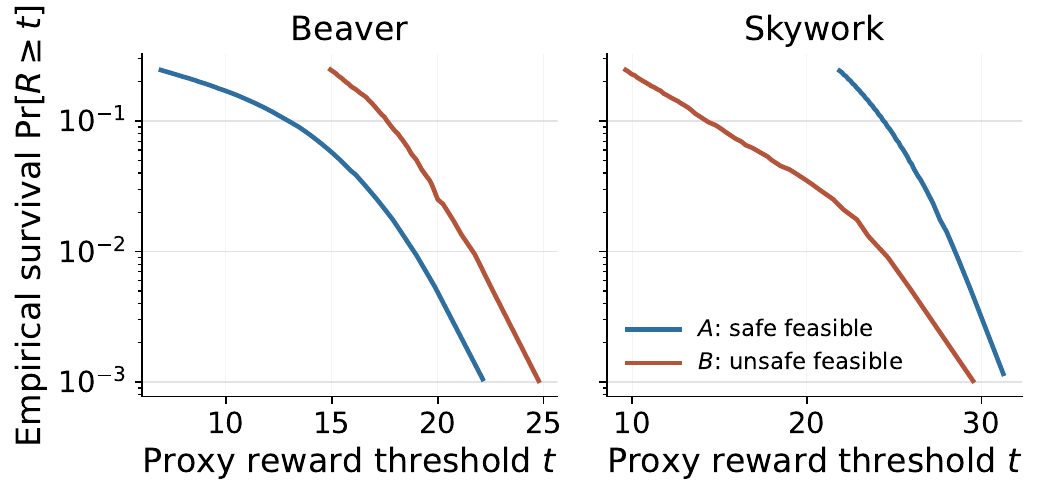}
    \caption{Conditional survival curves of learned reward scores among safe
    feasible ($A$) and unsafe feasible ($B$) responses. With Beaver, the
    unsafe feasible class has the heavier observed upper tail; with Skywork,
    the safe feasible class does. The panels use their respective proxy-score
    scales and should be interpreted within, rather than across, reward
    proxies.}
    \label{fig:skywork-tail-survival}
\end{figure*}

\Cref{fig:skywork-tail-survival} visualizes the conditional learned-score
survival curves within the safe feasible class $A(x,b)$ and the unsafe feasible
class $B(x,b)$. These curves are the score-distribution component of the joint
tails $\widehat{\Psi}_A$ and $\widehat{\Psi}_B$ in
\eqref{eq:joint-score-tail}; the class masses are held fixed across the Beaver
and Skywork ablations. With Beaver, unsafe feasible responses retain more mass
at high reward thresholds than safe feasible responses. With Skywork, the safe
feasible curve instead extends farther into the upper tail. This reversal is
consistent with the lower competitive unsafe-win term in
\Cref{fig:skywork-finite-n-decomposition} and with the finite-$N$ behavior in
\Cref{fig:skywork-hacking-scaling}. Because the figure estimates conditional
survival curves from a finite candidate pool, it is diagnostic evidence for the
tail mechanism rather than a verification of the asymptotic conditions in
\Cref{thm:hacking-asymptotic}.

\end{document}